%% file: main.tex
\documentclass[article,10pt]{wlscirep}
\usepackage[utf8]{inputenc}
\usepackage[T1]{fontenc}
\usepackage{amsmath,amssymb}
\usepackage{soul}
\usepackage{algorithm}
\usepackage{algpseudocode}
\usepackage{xcolor}
\usepackage{subcaption}

\title{Deep Kernel Learning of Dynamical Models from High-Dimensional Noisy Data}

\author[1,*]{Nicolò Botteghi}
\author[1]{Mengwu Guo}
\author[1]{Christoph Brune}
\affil[1]{Mathematics of Imaging and AI, University of Twente, Enschede, Netherlands}
\affil[*]{Corresponding author. E-mail: n.botteghi@utwente.nl}

\begin{abstract}

This work proposes a Stochastic Variational Deep Kernel Learning method for the data-driven discovery of low-dimensional dynamical models from high-dimensional noisy data.
The framework is composed of an encoder that compresses high-dimensional measurements into low-dimensional state variables, and a latent dynamical model for the state variables that predicts the system evolution over time. The training of the proposed model is carried out in an unsupervised manner, i.e., not relying on labeled data.
Our learning method is evaluated on the motion of a pendulum -- a well studied baseline for nonlinear model identification and control with continuous states and control inputs -- measured via high-dimensional noisy RGB images. Results show that the method can effectively denoise measurements, learn compact state representations and latent dynamical models, as well as identify and quantify modeling uncertainties. 

\end{abstract}
\begin{document}

\flushbottom
\maketitle

\thispagestyle{empty}

\section*{Introduction}

Understanding the evolution of dynamical systems over time by discovering their governing laws is essential for science and and engineering \cite{brunton2022data}. Traditionally, governing equations are derived from physical principles, such as conservation laws and symmetries. However, the governing laws are often difficult to unveil for many systems exhibiting strongly nonlinear behaviors. These complex behaviors are typically captured by high-dimensional noisy measurements, which makes it especially hard to identify the underlying principles. On the other hand, while measurement data are often abundant for many dynamical systems, physical equations, if known, may not exactly govern the actual system evolution due to various uncertainties. 

The progress of Machine Learning\cite{mitchell1997machine} and Deep Learning\cite{goodfellow2016deep}, combined with the availability of large amounts of data, has paved the road for new paradigms for the analysis and understanding of dynamical systems\cite{brunton2022data}. These new paradigms are not limited to the discovery of governing laws for system evolution, and have brought revolutionary advancements to the field of dynamical system control. In particular, Reinforcement Learning\cite{sutton2018reinforcement} (RL) has opened the door to model-free control directly from high-dimensional noisy measurements, in contrast to the traditional control techniques that rely on accurate physical models. RL has found its success in the nature-inspired learning paradigm through interaction with the world, in which the control law is solely a function of the measurements and learned by iteratively evaluating its performance a posteriori, i.e., after being applied to the system. Especially, RL stands outs in the control of complex dynamical systems \cite{arulkumaran2017deep}. However, RL algorithms may suffer from high computational cost and data inefficiency as a result of disregarding any prior knowledge about the world.

While data are often high-dimensional, many physical systems exhibit low-dimensional behaviors, effectively described by a limited number of latent state variables that can capture the principal properties of the systems. The process of encoding high-dimensional measurements into a low-dimensional latent space and extracting the predominant state variables is called, in the context of RL and Computer Science, State Representation Learning\cite{lesort2018state, botteghi2022unsupervised}. At the same time, its counterpart in Computational Science and Engineering is often referred to as Model Order Reduction\cite{quarteroni2014reduced}.

Reducing the data dimensionality and extracting the latent state variables is often the first step to explicitly represent a reduced model describing the system evolution. Due to their low dimensionality, such reduced models are often computationally lightweight and can be efficiently queried for making predictions of the dynamics\cite{hesthaven2022reduced} and for model-based control, e.g., Model Predictive Control \cite{camacho2013model} and Model-based RL \cite{sutton2018reinforcement}. The problem of dimensionality reduction and reduced-order modeling is traditionally tackled by the Singular Value Decomposition \cite{wall2003singular} (SVD) (depending on the context, the SVD is often referred to as Principal Component Analysis \cite{wold1987principal} or Proper Orthogonal Decomposition \cite{berkooz1993proper}).  Examples include the Dynamics Mode Decomposition \cite{schmid2010dynamic, proctor2016dynamic}, sparse identification of latent dynamics\cite{brunton2016discovering} (SINDy), operator inference \cite{ghattas2021learning, guo2022bayesian, peherstorfer2016data}, and Gaussian process surrogate modeling\cite{guo2019data}. More recently, Deep Learning \cite{goodfellow2016deep}, especially a specific type of neural network (NN) termed Autoencoder\cite{goodfellow2016deep} (AE), has been employed to learn compact state representations successfully. Unlike the SVD, an AE learns a nonlinear mapping from the high-dimensional data space to a low-dimensional latent space through an NN called encoder, as well as an inverse mapping through a decoder. AEs can be viewed as a nonlinear generalization of the SVD, enabling more powerful information compression and better expressivity. AEs have been used for manifold learning \cite{lee2020model}, in combination with SINDy for latent coordinate discovery \cite{champion2019data}, and in combination with NN-based surrogate models for latent representation learning towards control\cite{wahlstrom2015pixels, assael2015data}.


Whether we aim to identify parameters of a physical equation or learn the entire system evolution from data, we may face an unavoidable challenge stemming from data noise. Inferring complex dynamics from noisy data is not effortless, as the identification, understanding and quantification of various uncertainties is often required.
For example, uncertainties may derive from noise-corrupted sensor measurements, system parameters (e.g., uncertain mass, geometry, or initial conditions), modeling and/or approximation processes, and uncertain system behaviors that may be chaotic (e.g., in the motion of a double pendulum) or affected by unknown disturbances (e.g., uncertain external forces or inaccurate actuation).
When data-driven methods consider stochasticity and uncertainties quantification, AEs are often replaced with Variational AEs \cite{kingma2013auto} (VAEs) for learning low-dimensional states as probabilistic distributions. Samples from these distributions can be used for the construction of latent state models via Gaussian models\cite{fraccaro2017disentangled, krishnan2015deep,karl2016deep, buesing2018learning, doerr2018probabilistic} and nonlinear NN-based models\cite{hafner2019learning, hafner2019dream}. However, NN-based latent models often disregard the distinction among uncertainty sources, especially between the data noise in the measurements and the modeling uncertainties stemming from the learning process, and only estimate the overall uncertainty on the latent state space through the encoder of a VAE model. We argue, however, that disentangling the uncertainty sources is critical for identifying the governing laws and discovering the latent reduced-order dynamics. 

\begin{figure*}[h!]
    \centering
        \includegraphics[width=1.0\textwidth]{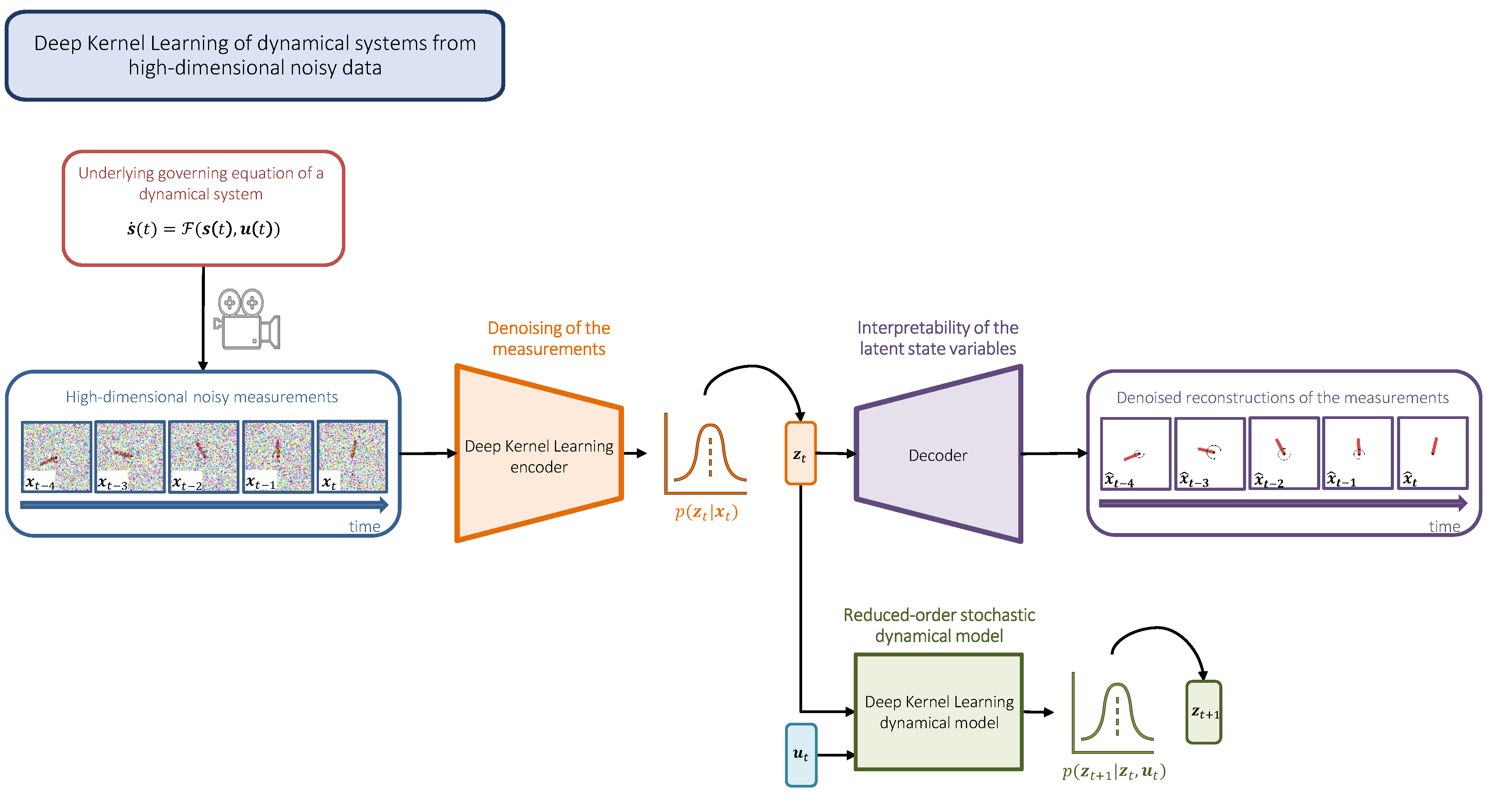}
        \caption{Deep Kernel Learning for data-driven dimensionality reduction, latent-state model learning, and uncertainty quantification of dynamical systems from high-dimensional noisy data.}
        \label{fig:DKL_framework}
\end{figure*}

In this work, we propose a data-driven framework for the dimensionality reduction, latent-state model learning, and uncertainty quantification based on high-dimensional noisy measurements generated by unknown dynamical systems (see Figure \ref{fig:DKL_framework}). In particular, we introduce a Deep Kernel Learning \cite{wilson2016deep} (DKL) encoder, which combines the highly expressive NN with a kernel-based probabilistic model of Gaussian process \cite{williams2006gaussian} (GP) to reduce the dimensionality and quantify the uncertainty in the noisy measurements simultaneously, followed by a DKL latent-state forward model that predicts the system dynamics with quantifiable modeling uncertainty, and an NN-based decoder designed to enable reconstruction, prevent representation-collapsing, and improve interpretability. Endowed with quantified uncertainties, such a widely applicable and computationally efficient method for manifold and latent model learning  is essential for data-driven physical modeling, control, and digital twinning. An implementation of our framework is available at: \url{https://github.com/nicob15/DeepKernelLearningOfDynamicalModels}.

\section*{Preliminaries}

In scalar-valued supervised learning, we have a set of $M$ $d$-dimensional input samples $\mathbf{X}=[\mathbf{x}_1, \dots, \mathbf{x}_M] \in \mathcal{X} \subset \mathbb{R}^d$ and the corresponding set of target data $\mathbf{y}=[{y}_1, \dots, {y}_M]^T$ $\in \mathcal{Y} \subset \mathbb{R}$ related by some unknown function $f^{\#}: \mathcal{X} \rightarrow \mathcal{Y}$.
The goal is to find a function $f$ that best approximates $f^{\#}$.  Many function approximators can be used to learn $f$, but here we introduce Gaussian process regression (GPR) \cite{williams2006gaussian} -- a non-parametric method for data-driven surrogate modeling and uncertainty quantification (UQ), {\color{black}deep NNs -- a popular class of parametric function approximators of Deep Learning}, and the Deep Kernel Learning\cite{wilson2016deep} (DKL) that combines the nonlinear expressive power of deep NNs with the advantages of kernel methods in UQ.  

\subsection*{Gaussian Process Regression}

A GP is a collection of random variables, any finite number of which follow a joint Gaussian distribution \cite{williams2006gaussian}.
\begin{equation}
    f(\mathbf{x}) \sim \mathrm{GP}(\mu(\mathbf{x}), k(\mathbf{x},\mathbf{x}';\mathbf{\gamma})), \ \ \ \ \ {y}=f(\mathbf{x}) + \epsilon, \ \ \ \ \ \epsilon \sim \mathcal{N(}0, \sigma_\epsilon^2)\,,
\end{equation}
where the GP is characterized by its mean function $\mu(\mathbf{x}) = \mathbb{E}[f(\mathbf{x})]$ and covariance/kernel function  $k(\mathbf{x},\mathbf{x}';\mathbf{\gamma})=k_{\mathbf{\gamma}}(\mathbf{x},\mathbf{x}')=\mathbb{E}[(f(\mathbf{x})-\mu(\mathbf{x}))(f(\mathbf{x}')-\mu(\mathbf{x}'))]$ hyperparameters $\gamma${\color{black}, $\mathbf{x}$ and $\mathbf{x}'$ being two input locations, and} $\epsilon$ is an independent added Gaussian noise term with variance $\sigma_\epsilon^2$. A popular choice of the kernel is the automatic relevance determination (ARD) squared exponential (SE) kernel:
\begin{equation}
    k_{\mathbf{\gamma}}(\mathbf{x}, \mathbf{x}')  = \sigma_f^2\exp\Big(-\frac{1}{2}\sum_{j=1}^d \frac{(x_j - x'_j )^2}{l_j^2} \Big)\,,
\end{equation}
where $\sigma_f$ is the standard deviation hyperparameter and $l_j$ ($1\leq j\leq d$) is the lengthscale along each individual input direction. The optimal values of GP hyperparameters $[\mathbf{\gamma},\sigma_\epsilon^2]=[\sigma_f^2, l_1, \dots, l_d,\sigma_\epsilon^2]$ can be estimated via maximum marginal likelihood given the training targets $\mathbf{y}$ \cite{williams2006gaussian}:
\begin{equation}
    [\mathbf{\gamma}^*,(\sigma_\epsilon^{2})^*] = \arg \max_{\mathbf{\gamma},\sigma_\epsilon^2}~\log p( \mathbf{y}|\mathbf{X})= \arg \max_{\mathbf{\gamma},\sigma_\epsilon^2} \left\{-\frac{1}{2} \mathbf{y}^T(k_{\mathbf{\gamma}}(\mathbf{X},\mathbf{X})+\sigma_\epsilon^2 \mathbf{I})^{-1}\mathbf{y} -\frac{1}{2} \log |k_{\mathbf{\gamma}}(\mathbf{X},\mathbf{X})+\sigma_\epsilon^2\mathbf{I}|-\frac{M}{2}\log(2\pi)\right\}\,.
    \label{eq:marginal_likelihood}
\end{equation}
Optimizing the GP hyperparameters through Equation (\ref{eq:marginal_likelihood}) requires to repeatedly inverse the covariance matrix $k_{\mathbf{\gamma}}(\mathbf{X},\mathbf{X})+\sigma_\epsilon^2\mathbf{I}$, which can be very expensive or even untrackable in the cases of high-dimensional inputs (e.g., images with thousands of pixels) or big datasets ($M\gg 1$).

Given the training data of input-output pairs $(\mathbf{X},\mathbf{y})$, the Bayes' rule gives a posterior Gaussian distribution of the noise-free outputs $\mathbf{f}^*$ at unseen test inputs $\mathbf{X}^*$:
\begin{equation}
\begin{split}
     \mathbf{f}^*| \mathbf{X}^*, \mathbf{X}, \mathbf{y} &\sim \mathcal{N}(\boldsymbol{\mu}^*, \boldsymbol{\Sigma}^*)\,, \\
     \boldsymbol{\mu}^*&=k_{\mathbf{\gamma}}(\mathbf{X}, \mathbf{X}^*)^T(k_{\mathbf{\gamma}}(\mathbf{X},\mathbf{X})+\sigma_\epsilon^2 \mathbf{I})^{-1}(\mathbf{y}-\mu(\mathbf{X}))\,,\\
      \boldsymbol{\Sigma}^*&=k_{\mathbf{\gamma}}(\mathbf{X}^*, \mathbf{X}^*) - k_{\mathbf{\gamma}}(\mathbf{X}, \mathbf{X}^*)^T(k_{\mathbf{\gamma}}(\mathbf{X},\mathbf{X})+\sigma_\epsilon^2 \mathbf{I})^{-1}k_{\mathbf{\gamma}}(\mathbf{X}, \mathbf{X}^*)\,.
\end{split}
\end{equation}

{\color{black}
\subsection*{Deep Neural Networks}

NNs are parametric universal function approximators\cite{chen1995universal} composed of multiple layers sequentially stacked together. Each layer contains a set of learnable parameters known as weights and biases. Collected in a vector $\theta$, these NN parameters are optimized via backpropagation\cite{goodfellow2016deep} for a function $f$ that best approximates $f^{\#}$:
\begin{equation}
    f(\mathbf{x}) = g(\mathbf{x}; \theta) 
\end{equation}
where $g(\mathbf{x};\theta)$ denotes an NN with input $\mathbf{x}$ and parameters $\theta$. } {\color{black} There are three prominent types of NN layers\cite{goodfellow2016deep}: fully-connected, convolutional, and recurrent. In practice, the three types of layers are often combined to deal with different characteristics of data and increase the expressivity of the NN model.}

\subsection*{Deep Kernel Learning}

To mitigate the limited scalability of GPs to high-dimensional inputs, often referred to as the curse of dimensionality, Deep Kernel Learning\cite{wilson2016deep, calandra2016manifold, bradshaw2017adversarial} was developed to exploit the nonlinear expressive power of deep NNs to learn compact data representations while maintaining the probabilistic features of kernel-based GP models for UQ. The key idea of DKL is to embed a deep NN, representing a nonlinear mapping from the data to the feature space, into the kernel function for GPR as follows: 
\begin{equation}
    k_\text{DKL}(\mathbf{x},\mathbf{x}';\mathbf{\gamma}, \mathbf{\theta}) = k_{\mathbf{\gamma}}(g(\mathbf{x};\mathbf{\theta}), g(\mathbf{x}';\mathbf{\theta}))\,,
\end{equation}
where $g(\mathbf{x};\mathbf{\theta})$ is an NN with input $\mathbf{x}$ and parameters (weights and biases) $\mathbf{\theta}$. Similar to conventional GPs, different kernel functions can be chosen. The GP hyperparameters and the NN parameters are jointly trained by maximizing the marginal likelihood as in Equation (\ref{eq:marginal_likelihood}). 

{\color{black} Thanks to its strong expressive power and versatility, DKL has gained attention in many fields of scientific computing, such as computer vision\cite{wilson2016deep, chen2016overview, ober2021promises}, natural language processing\cite{belanche2017bridging}, robotics\cite{calandra2016manifold}, and meta-learning\cite{tossou2019adaptive}.} However, DKL still suffers from computational inefficiency due to the need for repeatedly inverting the $M\times M$ covariance matrix in Equation \eqref{eq:marginal_likelihood} when the dataset is large ($M\gg 1$). In addition, the posterior will be intractable if we change to non-Gaussian likelihoods, and there is no efficient stochastic training\cite{goodfellow2016deep} (e.g., stochastic gradient descent) that is available for DKL models. All these facts make DKL unable to handle large datasets. To overcome these three limitations, Stochastic Variational DKL\cite{wilson2016stochastic} (SVDKL) was introduced. SVDKL utilizes variational inference\cite{williams2006gaussian} to approximate the posterior distribution with the best fitting Gaussian to a set of inducing data points sampled from the posterior. Our framework is built upon the SVDKL model.

Rather than other popular deep learning tools, SVDKL is chosen for three main reasons: (i) compared with deterministic NN-based models, GPs -- kernel-based models -- offer better quantification of uncertainties\cite{williams2006gaussian, wilson2016deep}, (ii) compared with Bayesian NNs\cite{kononenko1989bayesian}, SVDKL is computationally cheaper and feasible to the integration of any deep NN architecture, and (iii) compared with ensemble NNs, SVDKL is memory efficient as only a single model needs to be trained.

\section*{Methods}

In our work, we consider nonlinear dynamical systems generally written in the following form:
\begin{equation}
    \frac{d}{dt}\mathbf{s}(t) = \mathcal{F}(\mathbf{s}(t),\mathbf{u}(t)), \ \ \ \ \ \mathbf{s}(t_0)=\mathbf{s}_0, \ \ \ \ \ t\in[t_0,t_f] \,,
    \label{eq:dynamical_system}
\end{equation}
where $\mathbf{s}(t) \in \mathcal{S} \subset \mathbb{R}^n$ is the state vector at time $t$, $\mathbf{u}(t) \in \mathcal{U} \subset \mathbb{R}^m$ is the control input at time $t$, $\mathcal{F}:  \mathcal{S} \times  \mathcal{U} \rightarrow \mathcal{S}$ is a nonlinear function determining the evolution of the system given the current state $\mathbf{s}(t)$ and control input $\mathbf{u}(t)$, $\mathbf{s}_0$ is the initial condition, and $t_0$ and $t_f$ are the initial and final time,  respectively. In many real-world applications, the state $\mathbf{s}(t)$ is not directly accessible and the function $\mathcal{F}$ is unknown. In spite of this, we can obtain indirect information about the systems through measurements from different sensor devices (measurements can derive, for example, from cameras, laser scanners, or inertial measurement units). Due to the time-discrete nature of the measurements, we indicate with $\mathbf{x}_t$ the measurement vector at a generic time-step $t$, and $\mathbf{x}_{t+1}$ the measurement at time-step $t+1$.

Given a set of $M$ $d$-dimensional measurements $\mathbf{X}=[\mathbf{x}_1, \dots, \mathbf{x}_M] \in \mathcal{X} \subset \mathbb{R}^d$ with $d \gg 1$  and control inputs $\mathbf{U}=[\mathbf{u}_1, \dots, \mathbf{u}_{M-1}] \in \mathcal{U}$, we consider the problem of learning: (\emph{a}) a meaningful representation of the unknown states, and (\emph{b}) a surrogate model for $\mathcal{F}$. However, the high-dimensionality and noise corruption of measurement data makes the two-task learning problem extremely challenging.

\subsection*{Learning Latent State Representation from Measurements}

\begin{figure*}[h!]
    \centering
        \includegraphics[width=1.0\textwidth]{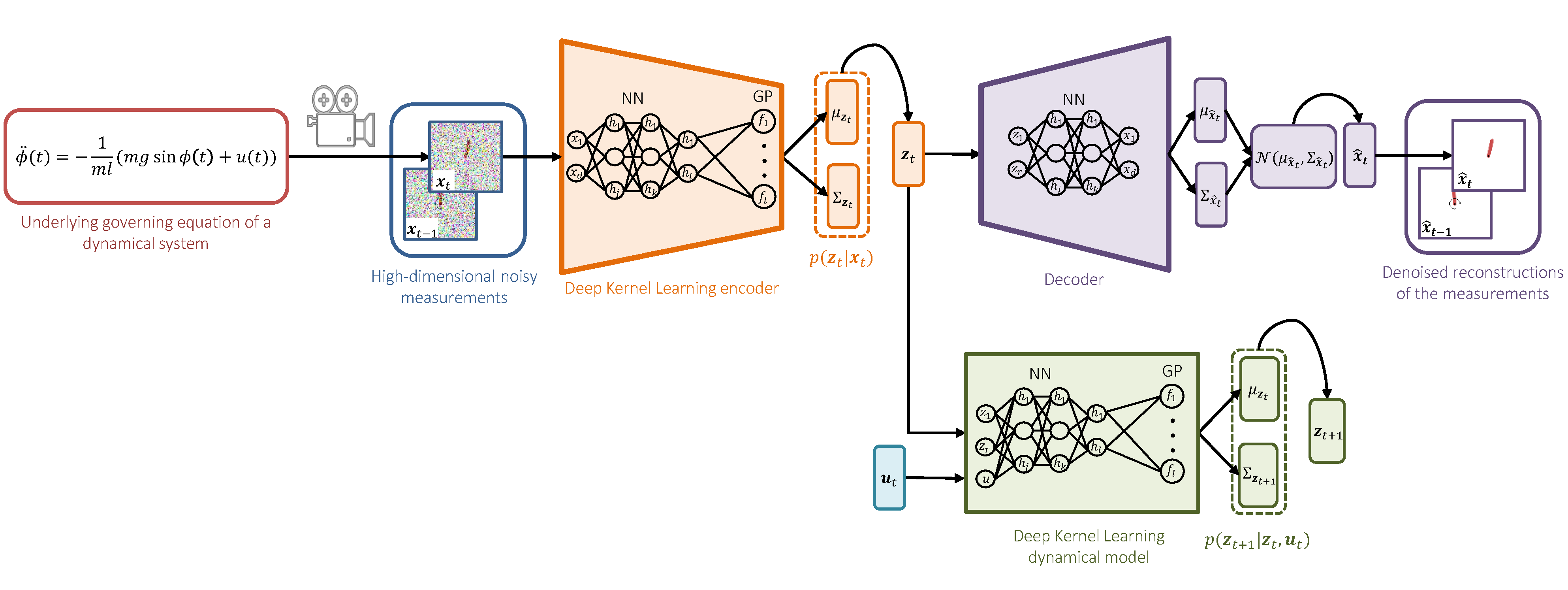}
        \caption{Uncertainty quantification and disentangling with Stochastic Variational Deep Kernel Learning for dynamical systems generating high-dimensional noisy data.}
        \label{fig:DKL_framework_method}
\end{figure*}

To begin with, we introduce an SVDKL encoder $E: \mathcal{X} \rightarrow \mathcal{Z}$ used to compress the measurements into a low-dimensional latent space $\mathcal{Z}$. Due to the measurement noise, rather than being deterministic, $E$ should map each measurement to a distribution over the latent state space $\mathcal{Z}$. The SVDKL encoder is depicted in Figure \ref{fig:DKL_framework_method}. A latent state sample can be obtained as:
\begin{equation}
\begin{split}
 z_{i,t} &= f_i^E(\mathbf{x}_t) + \epsilon_E,\quad \epsilon_E \sim \mathcal{N}(0, \sigma^2_E)\\
     f_i^E(\mathbf{x}_t) &\sim \mathrm{GP}(\mu(g_E(\mathbf{x}_t; \mathbf{\theta}_E)), k(g_E(\mathbf{x}_t; \mathbf{\theta}_E), g_E(\mathbf{x}'_{t'}; \mathbf{\theta}_E);\mathbf{\gamma}_{E,i})),\quad 1\leq i \leq |\mathbf{z}|\,,
\end{split}
\label{eq:posteriorE}
\end{equation}
where $z_{i,t}$ is the sample from the $i^{th}$ GP with kernel $k$ and mean $m$, $g_E(\mathbf{x}_t; \mathbf{\theta}_E)$ is the feature vector output of the NN part of the SVDKL encoder $E$, $\epsilon_E$ is an independently added noise, and $|\mathbf{z}|$ indicates the dimension of $\mathbf{z}$.

Because we have no access to the actual state values, we cannot directly use supervised learning techniques to optimize the parameters $[\mathbf{\theta}_E,\mathbf{\gamma}_E, \mathbf{\sigma}^2_E]$ of the SVDKL encoder. Therefore, we utilize a decoder neural network $D$ to reconstruct the measurements given the latent state samples. These reconstructions, denoted by $\hat{\mathbf{x}}_t$, are also used to generate trainable gradients for the SVDKL encoder, which is a common practice for {\color{black} training} VAEs. Similar {\color{black} to VAEs}, an important aspect of the architecture is the bottleneck created for the low dimensionality of the learned state space $\mathcal{Z}$. While the SVDKL encoder $E$ learns $p(\mathbf{z}_t|\mathbf{x}_t)$, the decoder $D$ learns the inverse mapping $p(\hat{\mathbf{x}}_t|\mathbf{z}_t)$ in which $\hat{\mathbf{x}}_t$ is the reconstruction of $\mathbf{x}_t$. We call this autoencoding architecture SVDKL-AE. To the best of our knowledge, this is the first attempt at training a DKL model without labeled data (unsupervisedly).

Given a randomly sampled minibatch of measurements, we can define the loss function for an SVDKL-AE as follows:
\begin{equation}
    \mathcal{L}_{E}(\mathbf{\theta}_E,\mathbf{\gamma}_E, \mathbf{\sigma}^2_E, \mathbb{\theta}_D) = \mathbb{E}_{\mathbf{x}_t \sim \mathbf{X}}[-\log p(\hat{\mathbf{x}}_t|\mathbf{z}_t)]\,,
\label{eq:loss_SVDKL_VAE}
\end{equation}
where $\hat{\mathbf{x}}_t|\mathbf{z}_t\sim\mathcal{N}(\mathbf{\mu}_{\hat{\mathbf{x}}_t},\boldsymbol{\Sigma}_{\hat{\mathbf{x}}_t})$ is obtained by decoding the samples of $\mathbf{z}_t|\mathbf{x}_t$ through $D$. By minimizing the loss function in Equation (\ref{eq:loss_SVDKL_VAE}) with respect to the encoder and decoder parameters, {\color{black} as analogously practiced with VAEs,} we can obtain a compact representation of the measurements. 

{\color{black} Though our SVDKL-AE resembles a VAE in terms of network architecture and training strategy, we highlight two major advantages of the SVDKL-AE, which have motivated its use in this work:
\begin{itemize}
    \item The SVDKL encoder explicitly learns the full distribution $p(\mathbf{z}|\mathbf{x})$ from which we can sample the latent states $\mathbf{z}$ reduced from the full-order states $\mathbf{x}$. A VAE only learns the mean vector and covariance matrix (often chosen to be diagonal) of an assumed joint Gaussian distribution. Clearly, SVDKL-AE should be able to deal with different types of complex distributions more effectively.
    \item SVDKL-AE can exploit the kernel structure of a Gaussian process to quantify uncertainties, even effectively in low-data regimes \cite{wilson2016stochastic, chen2016overview, belanche2017bridging}. The kernel choice can be tailored to incorporate prior knowledge into the data-driven modeling.
\end{itemize}}

\subsection*{Learning Latent Dynamical Model}
We aim to learn a surrogate dynamical model $F$ predicting the system evolution given the latent state variables sampled from $\mathcal{Z}$ and the control inputs in $\mathcal{U}$. Due to the uncertainties present in the system, we learn a stochastic model $F:\mathcal{Z} \times \mathcal{U} \rightarrow \mathcal{Z}$. Similar to $E$, the dynamical model $F$ is constructed using an SVDKL architecture. The next latent states $\mathbf{z}_{t+1}$ can be sampled with $F$:
\begin{equation}
\begin{split}
z_{i,t+1} &= f_i^F(\mathbf{z}_t,\mathbf{u}_t) + \epsilon_F,\quad \epsilon_F \sim \mathcal{N}(0, \sigma^2_F) \\
f_i^F(\mathbf{z}_t,\mathbf{u}_t) &\sim \mathrm{GP}(\mu(g_F(\mathbf{z}_t,\mathbf{u}_t; \mathbf{\theta}_F)), k(g_F(\mathbf{z}_t,\mathbf{u}_t; \mathbf{\theta}_F), g_F(\mathbf{z}'_{t'},\mathbf{u}'_{t'}; \mathbf{\theta}_F);\mathbf{\gamma}_{F,i})),\quad 1\leq i \leq |\mathbf{z}|\,,
\end{split}
\label{eq:posteriorF}  
\end{equation}
where $z_{i,t+1}$ is sampled from the $i^{th}$ GP, $g_F(\mathbf{z}_t,\mathbf{u}_t; \mathbf{\theta}_E)$ is the feature vector output of the NN part of the SVDKL dynamical model $F$, and $\epsilon_F$ is a noise term. 

Again, we do not have access to the true state values obtained by applying the (unknown) control law, but only the sequence of measurements at different time-steps. Here we employ a commonly used strategy in State Representation Learning\cite{hafner2019learning, hafner2019dream, hafner2020mastering}, which encodes the measurement $\mathbf{x}_{t+1}$ into the distribution $p(\mathbf{z}_{t+1}| \mathbf{x}_{t+1})$ through the SVDKL encoder $E$, and uses such a distribution as the target for $p(\mathbf{z}_{t+1}|\mathbf{z}_t, \mathbf{u}_t)$ given by the dynamical model $F$.   
Therefore, the dynamical model $F$ is trained by minimizing the Kullback-Leibler divergence between the distributions $p(\mathbf{z}_{t+1}| \mathbf{x}_{t+1})$ and $p(\mathbf{z}_{t+1}|\mathbf{z}_t, \mathbf{u}_t)$ (more details in Appendix). The loss for training $F$ is formulated as follows:
\begin{equation}
    \mathcal{L}_{\text{F}}(\mathbf{\theta}_F,\mathbf{\gamma}_F, \mathbf{\sigma}^2_F) = \mathbb{E}_{\mathbf{x}_t, \mathbf{x}_{t+1} \sim \mathbf{X}, \mathbf{u}_t \sim \mathbf{U}}[ \text{KL}[p(\mathbf{z}_{t+1}|\mathbf{x}_{t+1})||p(\mathbf{z}_{t+1}|\mathbf{z}_t,\mathbf{u}_t)]]]\,,
\label{eq:loss_SVDKL_F}
\end{equation}
where $p(\mathbf{z}_{t+1}|\mathbf{z}_t, \mathbf{u}_t)$ is obtained by feeding a sample from $p(\mathbf{z}_t|\mathbf{x}_t)$ and a control input $\mathbf{u}_t$ to $F$. 

\subsection*{Joint Training of Models}
Instead of training $E$ and $F$ separately, we train them jointly by allowing the gradients of the dynamical model $F$ to flow through the encoder $E$ as well. The overall loss function is 
\begin{equation}
    \mathcal{L}_{REP}(\mathbf{\theta}_E,\mathbf{\gamma}_E, \mathbf{\sigma}^2_E, \mathbf{\theta}_F,\mathbf{\gamma}_F, \mathbf{\sigma}^2_F, \mathbb{\theta}_D) = \mathbb{E}_{\mathbf{x}_t, \mathbf{x}_{t+1} \sim \mathbf{X}, \mathbf{u_t} \sim \mathbf{U}}[-\log p(\hat{\mathbf{x}}_t|\mathbf{z}_t))+\beta \text{KL}[p(\mathbf{z}_{t+1}|\mathbf{x}_{t+1})||p(\mathbf{z}_{t+1}|\mathbf{z}_t,\mathbf{u}_t)]]\,,
\label{eq:loss}
\end{equation}
in which $\beta=1.0$ is used to scale the contribution of the two loss terms.

\subsection*{Variational Inference}
The two SVDKL models in this work utilize variational inference to approximate the posterior distributions in \eqref{eq:posteriorE} and \eqref{eq:posteriorF} {\color{black} with a known family of candidate distributions (e.g., joint Gaussian distributions). The need for variational inference stems from the stochastic gradient descent optimization procedure used for the modeling training\cite{wilson2016stochastic}}. Therefore, we add two extra items to the loss function in Equation (\ref{eq:loss}), one for each SVDKL model in the following form:
\begin{equation}
    \mathcal{L}_{\text{var}}(\mathbf{\theta},\mathbf{\gamma}) = \text{KL}[p(\mathbf{v})||q(\mathbf{v})]\,,
\end{equation}
in which $p(\mathbf{v})$ is the posterior to be approximated over {\color{black}a collection of sampled locations $\mathbf{v}$ termed \emph{inducing points}}, and $q(\mathbf{v})$ represents an approximating candidate distribution. Similar to the original SVDKL work\cite{wilson2016stochastic}, the inducing points are placed on a grid.

\section*{Results}

\subsection*{Numerical Example}
\begin{figure*}[h!]
    \centering
        \includegraphics[width=0.75\textwidth]{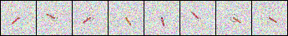}
        \caption{High-dimensional noisy measurements of the pendulum.}
        \label{fig:measurements_pendulum}
\end{figure*}

For our experiments, we consider the pendulum described by the following equation:
\begin{equation}
    \ddot{\phi}(t) = - \frac{1}{ml}(mg \sin{\phi}(t)+u(t))\,,
    \label{eq:pendulum_dynaimcs}
\end{equation}
where $\phi$ is the angle of the pendulum, $\ddot{\phi}$ is the angular acceleration, $m$ is the mass, $l$ is the length, and $g$ denotes the gravity acceleration. We assume no access to $\phi$ or its derivatives, and the measurements are RGB images of size $84 \times 84 \times 3$ obtained through an RGB camera. Examples of high-dimensional and noisy measurements are shown in Figure \ref{fig:measurements_pendulum}.
The measurements are collected by applying torque values $u$ sampled from a random control law with different initial conditions. The training set is composed of 15000 data tuples $(\mathbf{x}_t,\mathbf{u}_t, \mathbf{x}_{t+1})$, while the test set is composed of 2000 data tuples. Different random seeds are used for collecting training and test sets. The complete list of hyperparameters used in our experiments is shown in Appendix. 

\subsection*{Denoising}

\begin{figure*}[h!]
    \centering
        \includegraphics[width=1.0\textwidth]{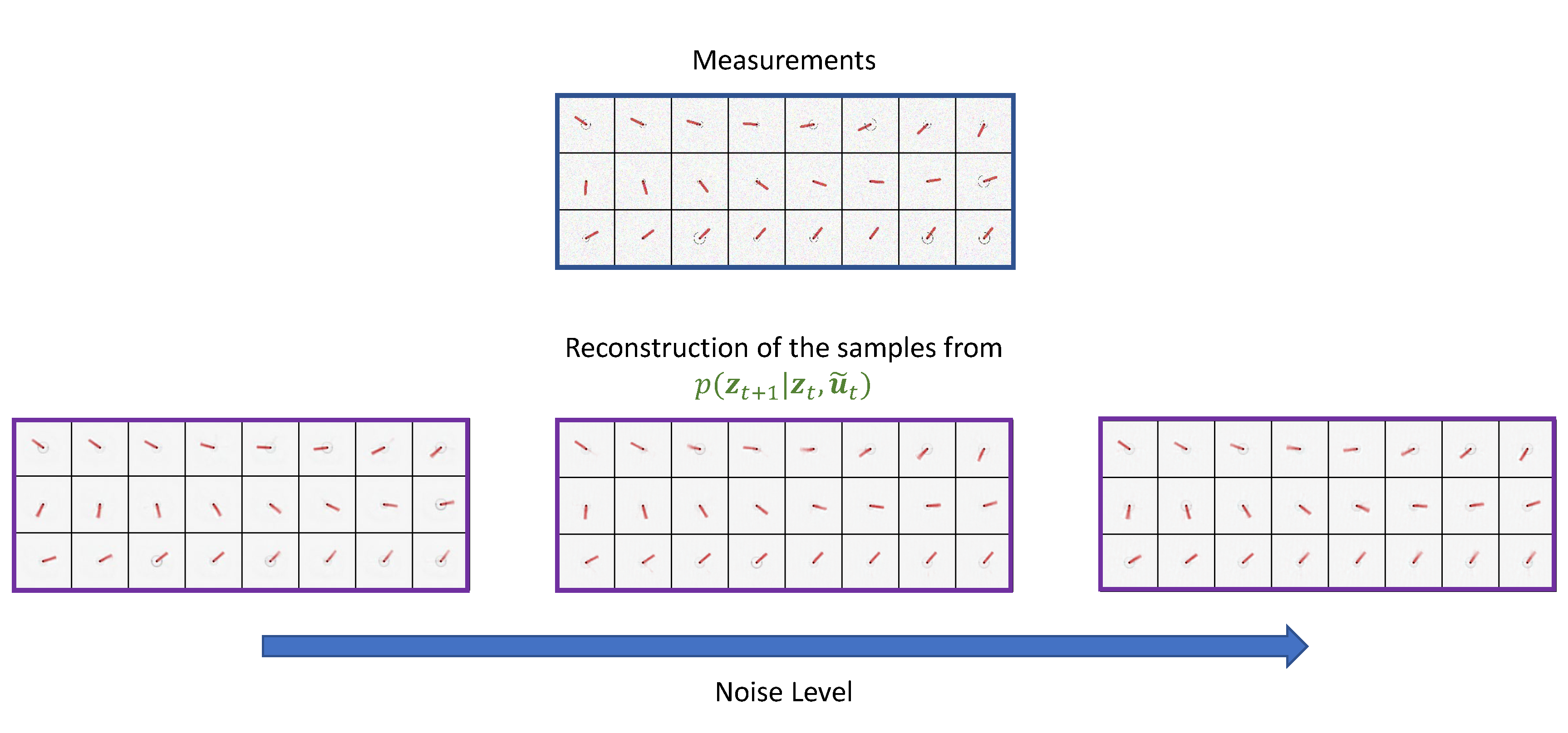}
        \caption{Reconstructions $\hat{\mathbf{x}}_{t+1}$ with different noise levels in the control inputs $\mathbf{u}_t$. {\color{black}As shown by the sharp reconstructions of $\mathbf{z}_{t+1}$, the SVDKL forward model can denoise the corrupted control inputs $\mathbf{u}_t$ and predict the dynamics accurately.} }
        \label{fig:denoising_u}
\end{figure*}

\begin{figure*}[h!]
    \centering
        \includegraphics[width=0.75\textwidth]{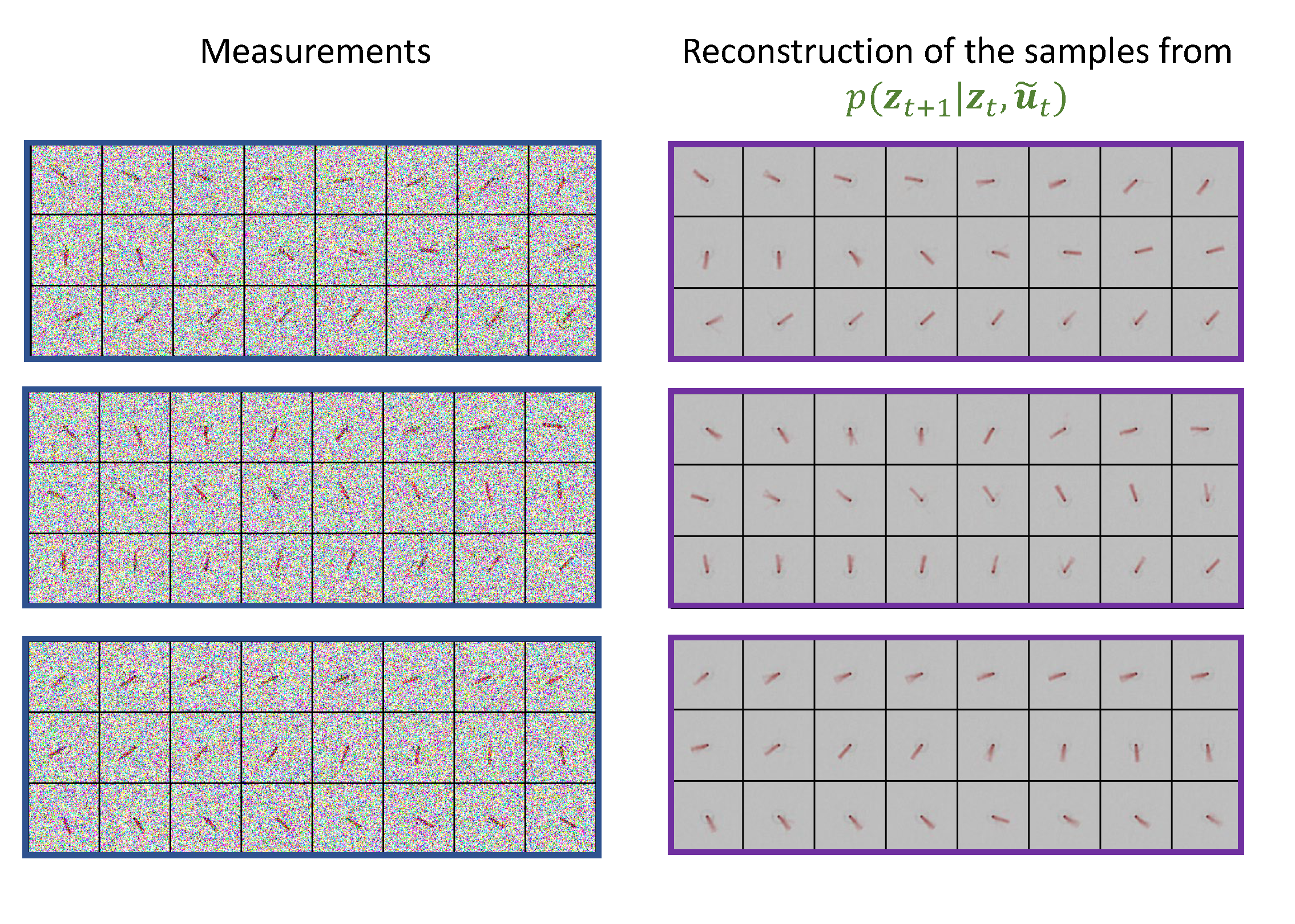}
        \caption{Reconstructions $\hat{\mathbf{x}}_{t+1}$ with different noise levels in both the measurements $\Tilde{\mathbf{x}}_t$ and the control inputs $\Tilde{\mathbf{u}}_t$. {\color{black}With both the measurements and control inputs corrupted by significant noise, the proposed model presents good performance in denoising.}}
        \label{fig:denoising_xu}
\end{figure*}

\begin{figure*}[h!]
    \centering
        \includegraphics[width=1.0\textwidth]{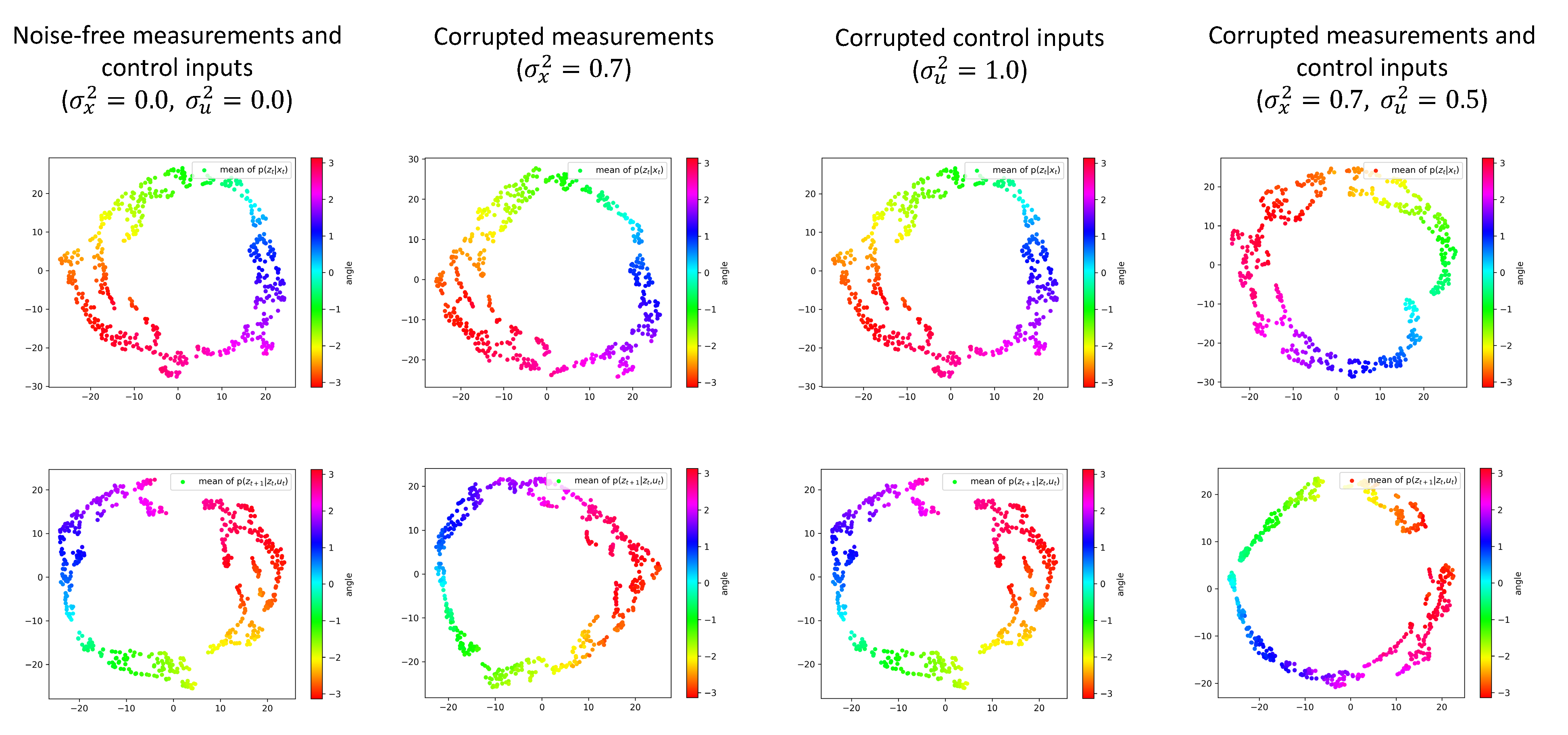}
        \caption{t-SNE visualization for the means of current {\color{black}(top)} and next {\color{black}(bottom)} latent state distributions with different noise levels in the measurements and control inputs. The color bar represents the true angle of the pendulum. {\color{black} As expected for a good denoising scheme, the change in the means of latent states is inconsiderable while the level of noise in the measurements and control inputs is increased significantly.} }
        \label{fig:mean_representations}
\end{figure*}

In Figure \ref{fig:denoising_x}, we show the denoising capability of the proposed framework by visualizing the reconstructions of the high-dimensional noisy measurements. The measurements are corrupted by additive Gaussian noise $\mathcal{N}(0, \sigma^2_x)$:
\begin{equation}
    \Tilde{\mathbf{x}}_t = \mathbf{x}_t + \epsilon_{\mathbf{x}}, \quad \epsilon_{\mathbf{x}} \sim \mathcal{N}(0, \sigma^2_x)\,.
\end{equation}
\begin{figure*}[h!]
    \centering
        \includegraphics[width=0.75\textwidth]{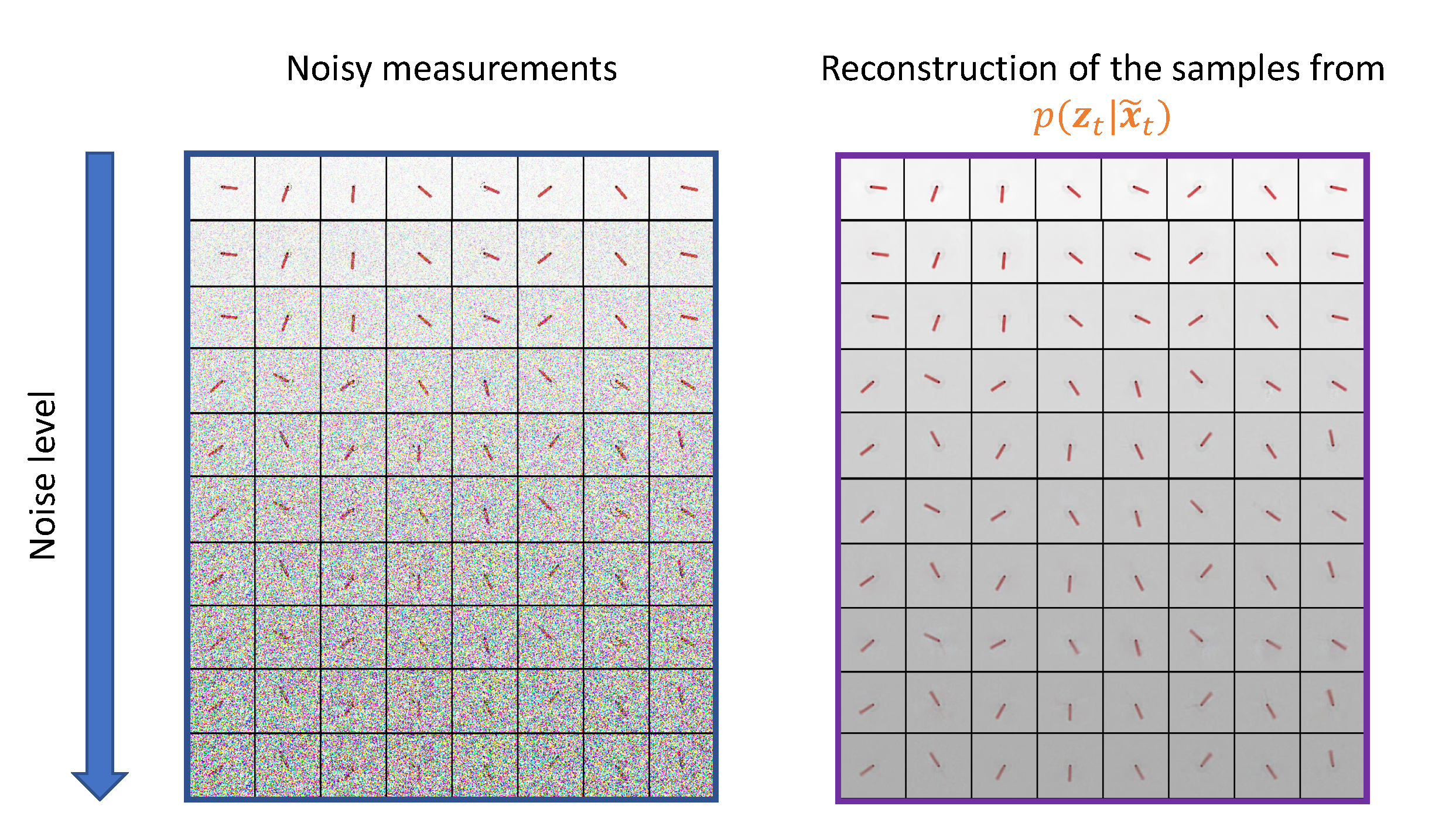}
        \caption{Reconstructions $\hat{\mathbf{x}}_{t}$  with different noise levels in the measurements $\Tilde{\mathbf{x}}_t$. {\color{black}As shown by the sharp reconstructions of $\mathbf{z}_t$, SVDKL-AE can effectively denoise the measurements.}}
        \label{fig:denoising_x}
\end{figure*}
Moreover, in Figure \ref{fig:denoising_u}, we show the reconstructions of the next latent states $\mathbf{z}_{t+1}$ sampled from the dynamic model distribution $p(\mathbf{z}_{t+1}|\mathbf{z}_t, \mathbf{u}_t)$ when the control input $\mathbf{u}_t$ is corrupted by Gaussian noise $\mathcal{N}(0, \sigma^2_u)$:
\begin{equation}
    \Tilde{\mathbf{u}}_t = \mathbf{u}_t + \epsilon_{\mathbf{u}}, \quad \epsilon_{\mathbf{u}} \sim \mathcal{N}(0, \sigma^2_u)\,.
\end{equation}
Eventually, in Figure \ref{fig:denoising_xu}, we show the reconstructions when $\mathbf{x}_t$ and $\mathbf{u}_t$ are simultaneously corrupted by Gaussian noises $\mathcal{N}(0, \sigma^2_x)$ and $\mathcal{N}(0, \sigma^2_u)$, respectively.
In all the three cases, our framework can properly denoise the input measurements by encoding the predominant features into the latent space. To support this claim, we show, in Figure \ref{fig:mean_representations}, the means of the current and next latent state distributions with  different noise corruptions. It is worth noting that the means of such distributions are a high-quality representation of the actual dynamics of the pendulum. 
Due to the dimensionality ($>2$) of the latent state space, we use t-SNE\cite{van2008visualizing} to visualize the results in 2-dimensional figures with the color bar representing the actual angle of the pendulum. The smooth change of the representation with respect to the true angle indicates its high quality. {\color{black}Moreover, it is worth mentioning that, as the level of noise in the measurements and control inputs is increased dramatically, the changes in the means of learned distributions are insignificant because of the denoising capability of the proposed model.} 

\subsection*{Prediction of Dynamics}

To better demonstrate how well the framework performs in prediction under uncertainties, we modify the pendulum dynamics in Equation \eqref{eq:pendulum_dynaimcs} to account for stochasticity due to, for example, external disturbances:
\begin{equation}
    \ddot{\phi}(t) = - \frac{1}{ml}(mg \sin{\phi}(t)+u(t)+\epsilon_{dyn}), \quad \epsilon_{dyn} \sim \mathcal{N}(0, \sigma^2_{dyn}).
\end{equation}
Again, we include an independently added Gaussian noise. While $\epsilon_\mathbf{x}$ and $\epsilon_\mathbf{u}$ are noise terms added to the noise-free measurements $\mathbf{x}_t$ and control inputs $\mathbf{u}_t$ to model, for example, the noise deriving from the sensor devices, $\epsilon_{dyn}$ approximates an unknown disturbance on the actual pendulum dynamics.

\begin{figure*}[h!]
    \centering
        \includegraphics[width=1.0\textwidth]{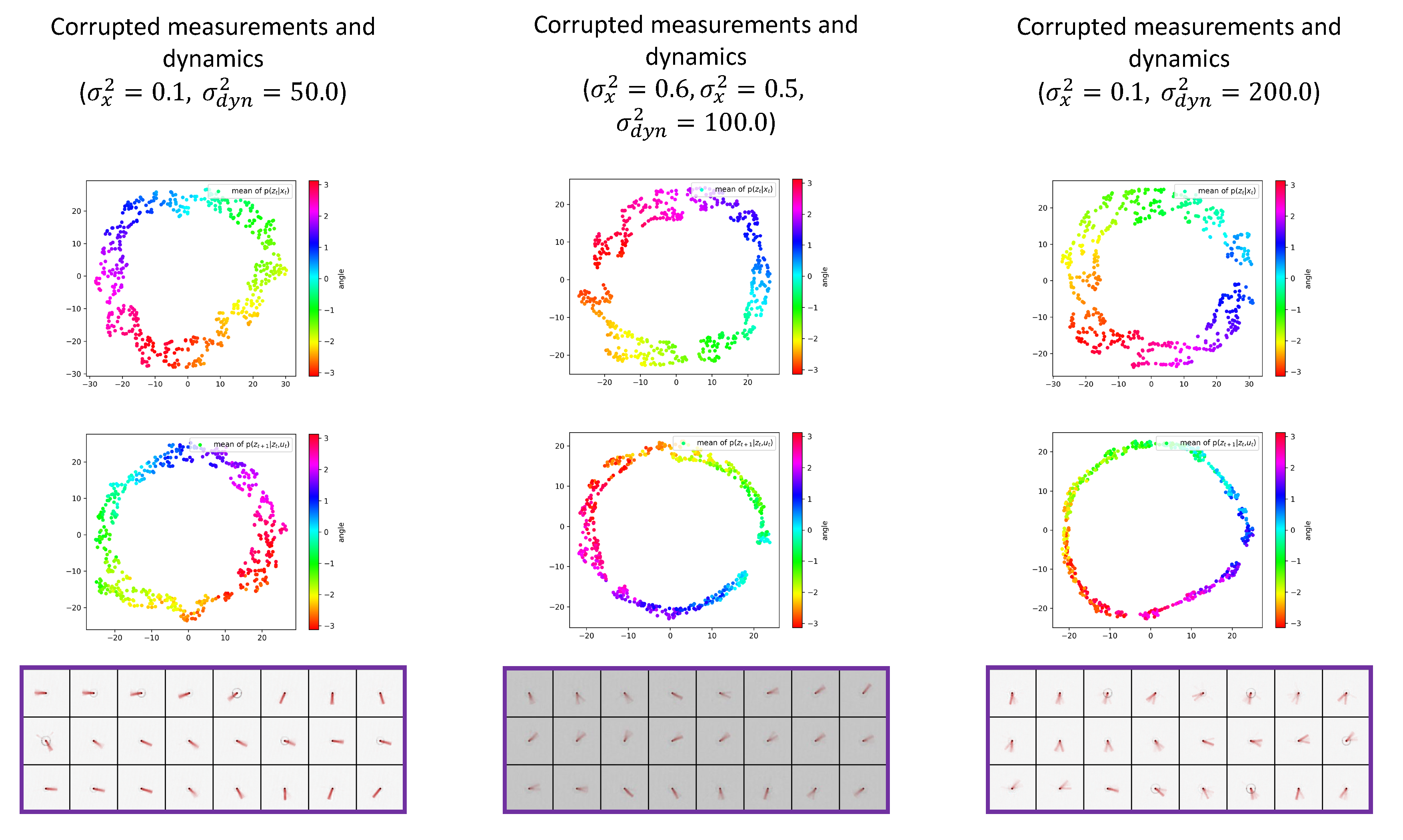}
        \caption{t-SNE visualization for the means of current and next latent state distributions with different levels of measurement noise and dynamics stochasticity, and the corresponding reconstructions $\hat{\mathbf{x}}_{t+1}$.}
        \label{fig:stoch_dynamics}
\end{figure*}

In Figure \ref{fig:stoch_dynamics}, we show the means of $p(\mathbf{z}_t|\mathbf{x}_t)$ and $p(\mathbf{z}_{t+1}|\mathbf{z}_t, \mathbf{u}_t)$ with different noise levels, and the corresponding (decoded) reconstructions of $\mathbf{z}_{t+1}$ samples from $p(\mathbf{z}_{t+1}|\mathbf{z}_t, \mathbf{u}_t)$. From the mean of $p(\mathbf{z}_t|\mathbf{x}_t)$, we can notice that the SVDKL encoder properly denoises the measurements and extracts the latent state variables when both measurement noise and disturbance on the actual pendulum dynamics exist. The SVDKL dynamical model recovers the mean of $p(\mathbf{z}_{t+1}|\mathbf{z}_t, \mathbf{u}_t)$ when the dynamical evolution of the pendulum is affected by an unknown stochastic disturbance. Even with high level of disturbance, though the system evolution over time becomes stochastic and more difficult to predict, $p(\mathbf{z}_{t+1}|\mathbf{z}_t, \mathbf{u}_t)$ can still capture and predict the evolution. Eventually, we can visualize the overall uncertainty in the dynamics reflected by the reconstruction of $\mathbf{z}_{t+1}$ via the decoder $D$. Note that the reconstructions in Figure \ref{fig:stoch_dynamics} are obtained by averaging 10 independent samples per data point.



{\color{black}
\subsection*{Uncertainty Quantification}

In this subsection, we show that the proposed SVDKL-AE enables the quantification of uncertainties in model predictions. It is worth mentioning that visualizing UQ properly is a  commonly recognized challenging task in unsupervised learning. 
The learned latent state vector $\mathbf{z}$ is 20-dimensional. To visualize the UQ capability of the proposed model, we select the $i$th-component ($i=12$ and $13$ in Figures \ref{fig:UQ_meas_noise=0.0-0.7} and \ref{fig:UQ_meas_noise=0.5-0.5}, respectively) of the state vector that is correlated with the physical states, and depict its predictive uncertainty bounds  for different noise levels ($\sigma_x^2=0.0, \sigma_u^2=0.7$ and $\sigma_x^2=0.5, \sigma_u^2=0.5$, rspectively). Because we are investigating an unsupervised learning problem, the latent variables may not have a direct physical interpretation. However, a good latent representation should present strong correlation with the physical states, and a proper UQ should reflect the existence of noise in the measurements $\mathbf{x}$ and/or control inputs $\mathbf{u}$. 
\begin{figure}[h!]
     \centering
     \begin{subfigure}[b]{0.49\textwidth}
         \centering
         \includegraphics[width=1.0\textwidth]{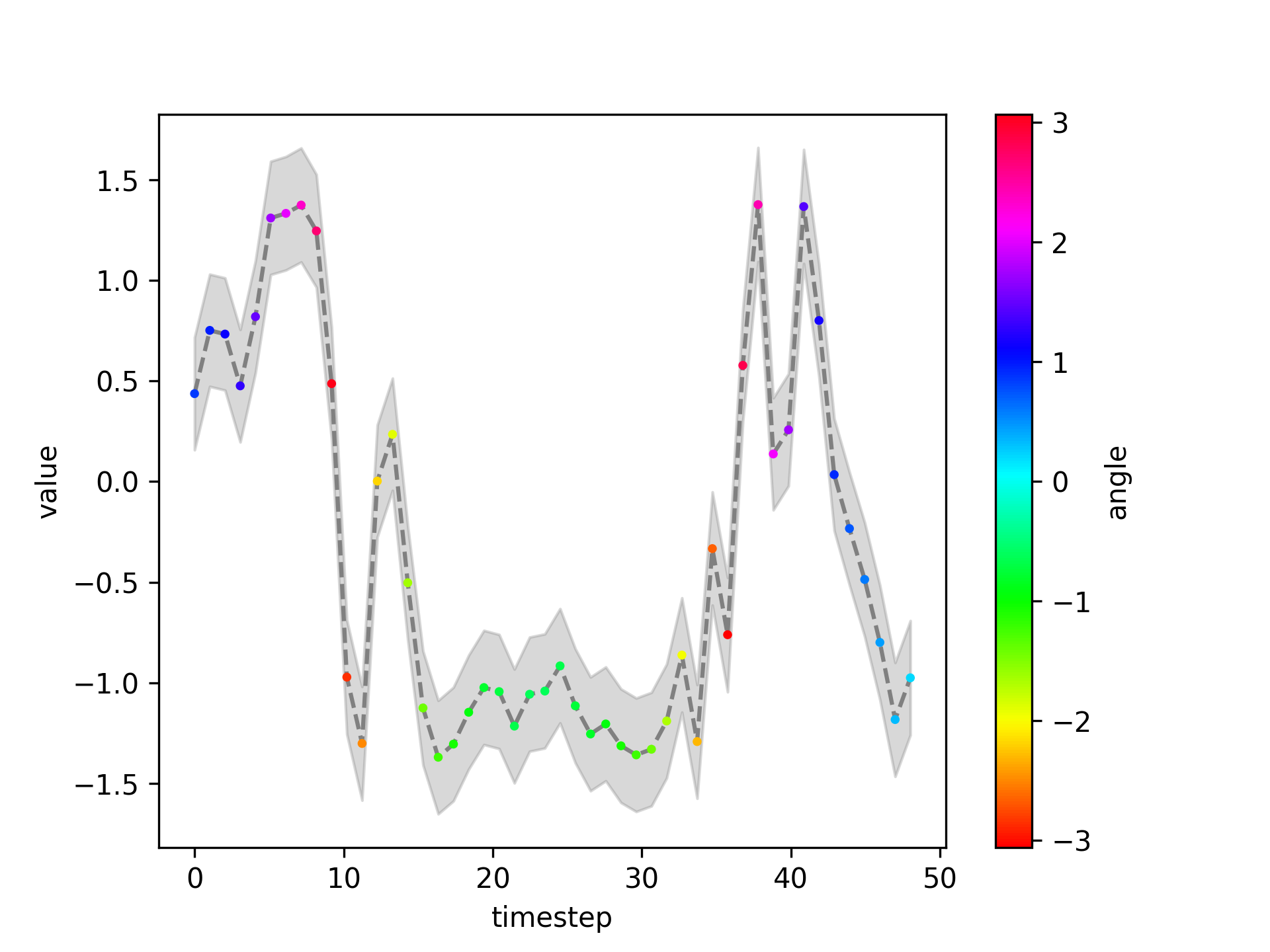}
         \caption{}
         \label{fig:z_pendulum}
     \end{subfigure}
     \hfill
     \begin{subfigure}[b]{0.49\textwidth}
         \centering
         \includegraphics[width=1.0\textwidth]{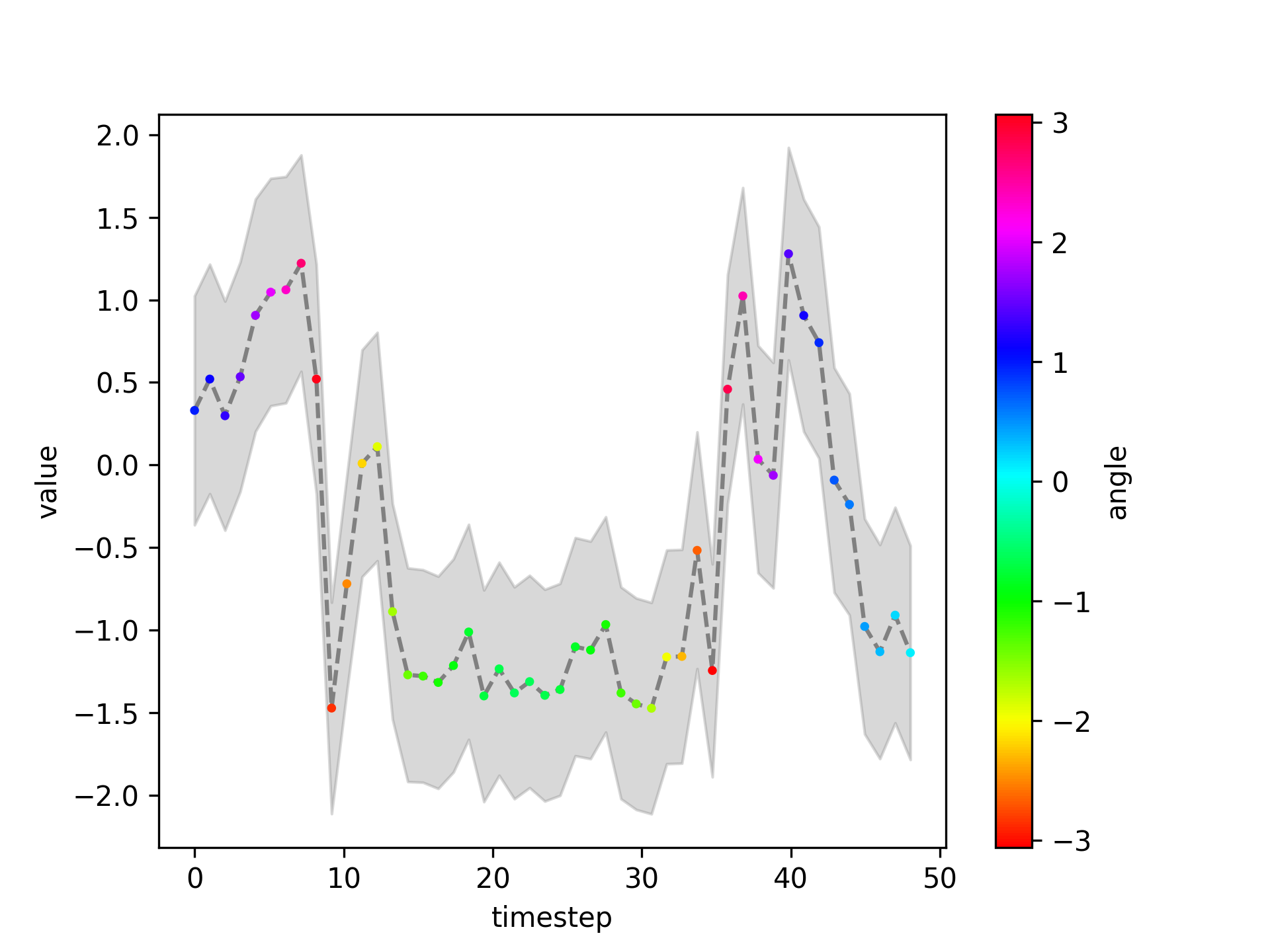}
         \caption{}
         \label{fig:z_next_pendulum}
     \end{subfigure}
     \hfill
    \begin{subfigure}[b]{0.49\textwidth}
         \centering
         \includegraphics[width=1.0\textwidth]{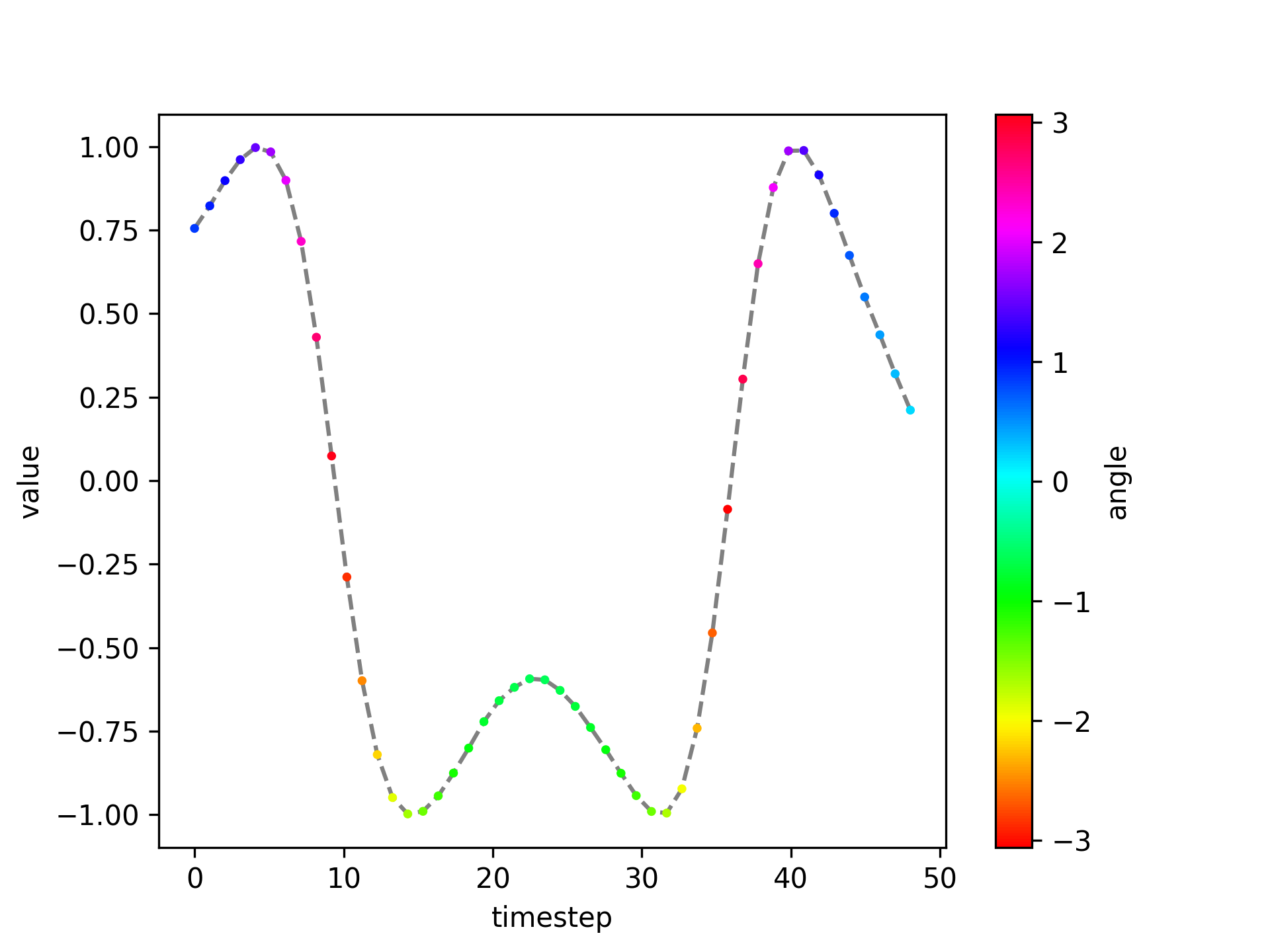}
         \caption{}
         \label{fig:true_state_var}
     \end{subfigure}
     \begin{subfigure}[b]{0.49\textwidth}
         \centering
         \includegraphics[width=1.0\textwidth]{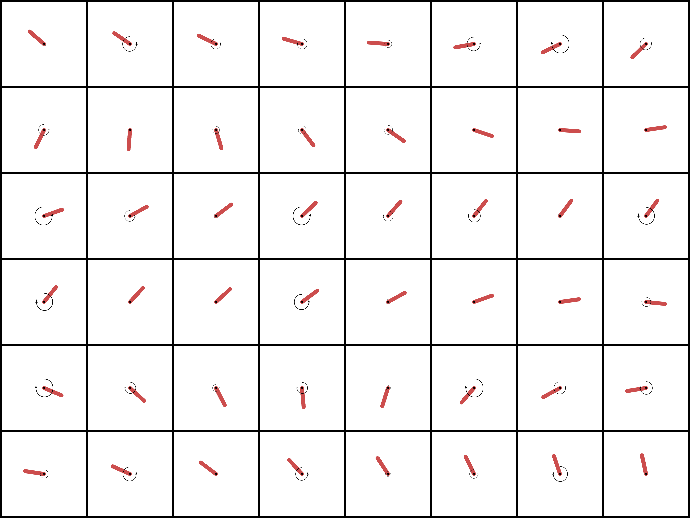}
         \caption{}
         \label{fig:traj_meas_noise=0.0-0.7}
     \end{subfigure}
        \caption{Estimated uncertainties over the 12th-components of the latent state vectors $\mathbf{z}_t$ (Figure \ref{fig:z_pendulum}) and $\mathbf{z}_{t+1}$ (Figure \ref{fig:z_next_pendulum}) predicted by the proposed model for a sampled trajectory. The $y$-position of the pendulum is given in Figure \ref{fig:true_state_var}, and the corresponding high-dimensional noisy measurements ($\sigma_x^2=0.0, \sigma_u^2=0.7$) are shown in Figure \ref{fig:traj_meas_noise=0.0-0.7}. The uncertainty bands are given by $\pm$ one standard deviation in the predictive distributions.}
        \label{fig:UQ_meas_noise=0.0-0.7}
\end{figure}
\begin{figure}[h!]
     \centering
     \begin{subfigure}[b]{0.49\textwidth}
         \centering
         \includegraphics[width=1.0\textwidth]{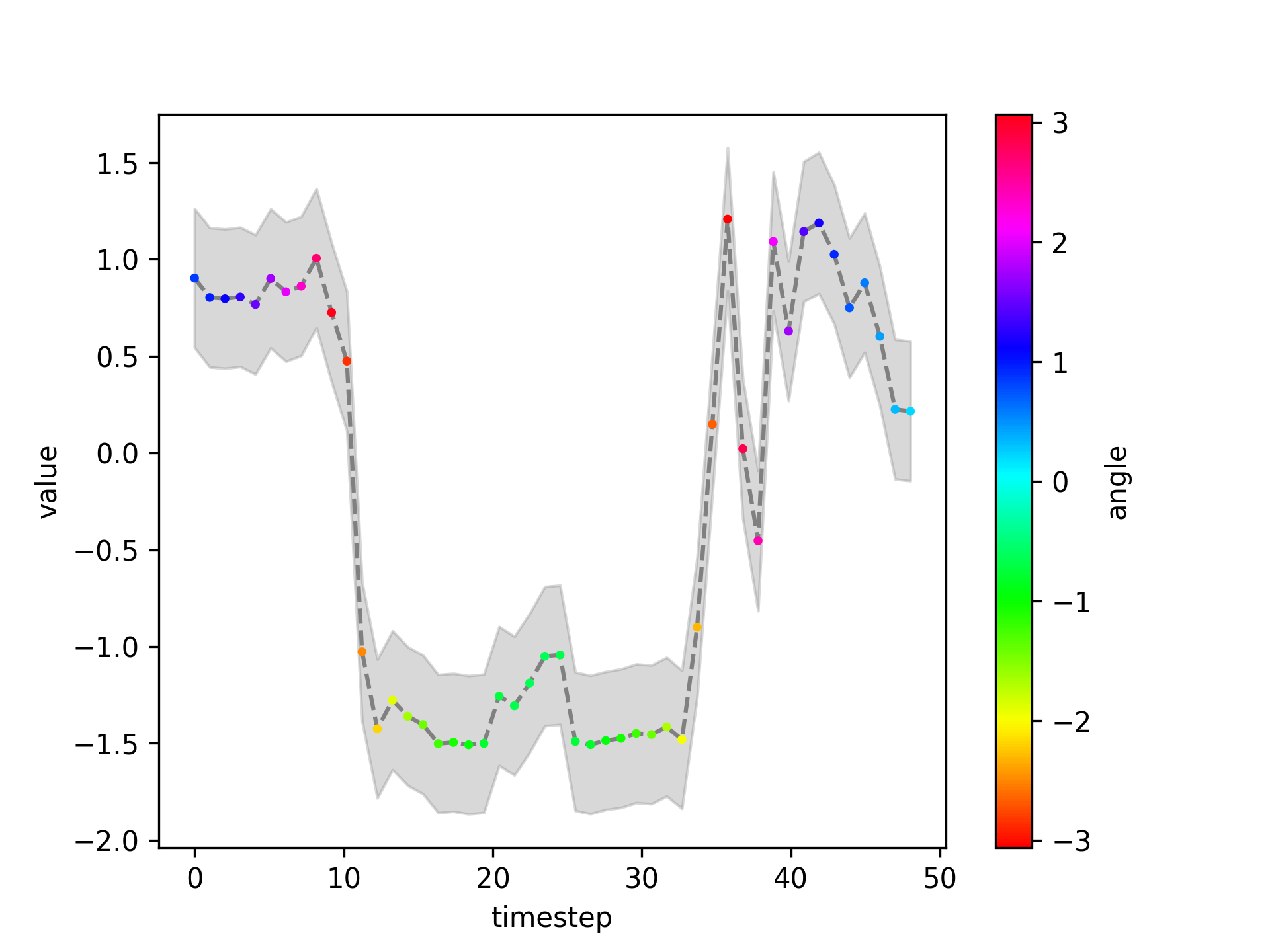}
         \caption{}
         \label{fig:z_pendulum_0.5-0.5}
     \end{subfigure}
     \hfill
     \begin{subfigure}[b]{0.49\textwidth}
         \centering
         \includegraphics[width=1.0\textwidth]{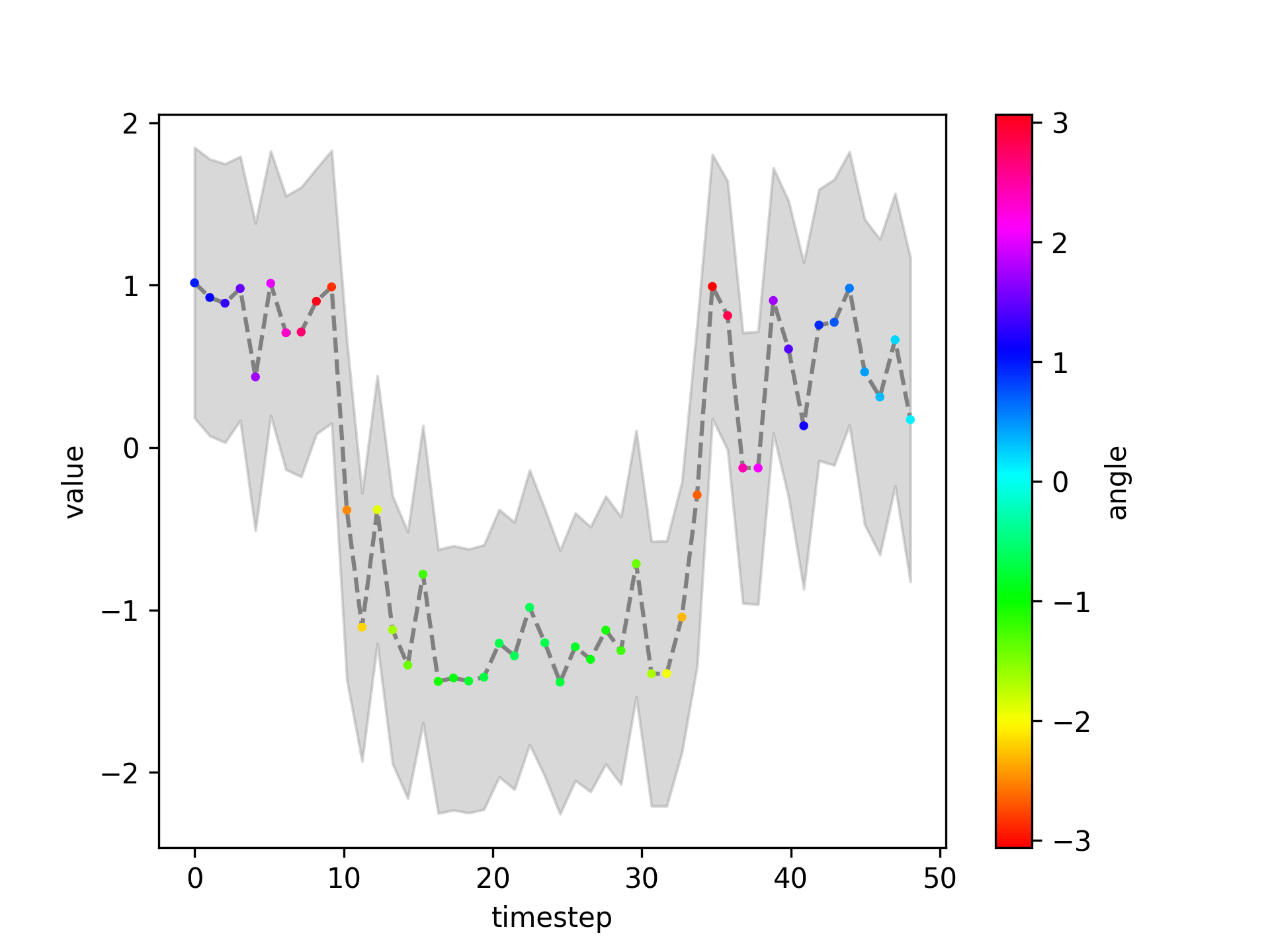}
         \caption{}
         \label{fig:z_next_pendulum_0.5-0.5}
     \end{subfigure}
     \begin{subfigure}[b]{0.49\textwidth}
         \centering
         \includegraphics[width=1.0\textwidth]{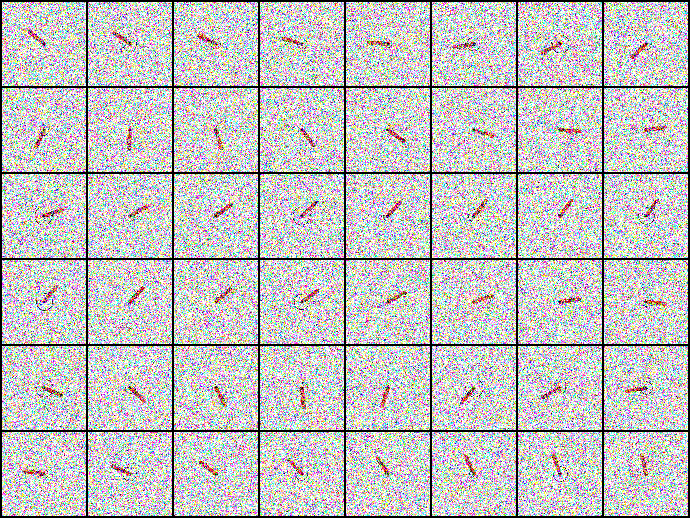}
         \caption{}
         \label{fig:traj_meas_noise=0.5-0.5}
     \end{subfigure}
        \caption{Estimated uncertainties over the 13th-components of the latent state vectors $\mathbf{z}_t$ (Figure \ref{fig:z_pendulum_0.5-0.5}) and $\mathbf{z}_{t+1}$ (Figure \ref{fig:z_next_pendulum_0.5-0.5}) predicted by the proposed model for a sampled trajectory. The corresponding high-dimensional noisy measurements ($\sigma_x^2=0.5, \sigma_u^2=0.5$) are shown in Figure \ref{fig:traj_meas_noise=0.5-0.5}. The uncertainty bands are given by $\pm$ one standard deviation in the predictive distributions.}
        \label{fig:UQ_meas_noise=0.5-0.5}
\end{figure}
As seen in the figures, the proposed framework achieves a good system representation with latent variables that are highly correlated with physical quantities of interest (see Figures \ref{fig:z_pendulum}-\ref{fig:true_state_var} and \ref{fig:z_pendulum_0.5-0.5}-\ref{fig:z_next_pendulum_0.5-0.5}), 
devised by uncertainty bands reflecting data noise and modeling errors (see Figures \ref{fig:z_pendulum}-\ref{fig:z_next_pendulum} in comparison with \ref{fig:z_pendulum_0.5-0.5}-\ref{fig:z_next_pendulum_0.5-0.5}.)
}

\section*{Discussion and Future Work}

Though well researched in supervised learning\cite{abdar2021review}, uncertainty quantification is still an understudied topic in unsupervised dimensionality reduction and latent model learning. However, the combination of these two tasks has the potential to open new doors to the discovery of governing principles of complex dynamical systems from high-dimensional noisy data. Our proposed method provides convincing indications that combining deep NNs with kernel-based models is promising for the analysis of high-dimensional noisy data. Our general framework relies only on the observations of measurements and control inputs, making it applicable to all physical modeling, digital twinning, weather forecast, and patient-specific medical analysis. 

Learning compact state representations and latent dynamical models from high-dimensional noisy observations is a critical element of Optimal Control and Model-based RL. In both, the disentanglement of measurement and modeling uncertainties will play a crucial role in optimizing control laws, as well as in devising efficient exploration of the latent state space to aid the collection of new, informative samples for model improvement. The quantified uncertainties can be exploited for Active Learning\cite{settles2012active} to steer the data sampling\cite{fasel2022ensemble}. 

\section*{Conclusions}
SVDKL models are integrated into a novel general workflow of unsupervised dimensionality reduction and latent dynamics learning, combining the expressive power of deep NNs with the uncertainty quantification abilities of GPs. The proposed method has shown good capability of generating interpretable latent representations and denoised reconstructions of high-dimensional, noise-corrupted measurements, see Figures \ref{fig:mean_representations} and \ref{fig:denoising_x}-\ref{fig:denoising_xu}, respectively. It has also been demonstrated that this method can deal with stochastic dynamical systems by identifying the source of stochasticity.

\section*{Acknowledgements}

The second author acknowledges the support from Sectorplan Bèta (NL) under the focus area \emph{Mathematics of Computational Science}. 

\section*{Author Contributions Statement}

N.B. and M.G. conceived the mathematical models, N.B. implemented the methods and designed the numerical experiments, N.B. and M.G. interpreted the results, N.B. wrote the first draft, and M.G. and C.B. reviewed the manuscript. All authors gave approval for the final manuscript.

\section*{Data Availability}
The datasets used and/or analysed during the current study are available from the corresponding author on reasonable request.

\section*{Additional Information}

The authors declare no competing interests.



\appendix
\input{appendixA}

\bibliography{sample}

\end{document}

%% file: appendixA.tex
\appendix
\section*{Appendix A Implementation Details}\label{appendix_A}

The proposed SVDKL framework is constructed using the GPyTorch \cite{gardner2018gpytorch} library. An implementation of our framework is available at: \url{https://github.com/nicob15/DeepKernelLearningOfDynamicalModels}.

\subsection*{A.1 SVDKL-AE Architecture}\label{svdkl-ae}

The input measurements are $84\times84\times3$ RGB images. To allow the encoder to learn the angle of the pendulum and its angular velocity, two consecutive frames are stacked together, making the input measurements of size $84\times84\times6$. The SVDKL encoder $E$ is composed of 4 convolutional layers with 32 filters per layer. The convolutional filters are of size $(3\times3)$ and shifted across the images with stride 1 (only the first convolutional layer has stride 2 to quickly reduce the input dimensionality). Batch normalization is also used after the 2nd and 4th convolutional layers. A similar convolutional architecture is used in PlaNet\cite{hafner2019learning} and Dreamer\cite{hafner2019dream}. The output features of the last convolutional layer are flattened and fed to two final fully connected layers of dimensions 256 and 20, respectively, compressing the features to a 20-dimensional feature vector. Each layer has ELU activations, except the last fully-connected layer with a linear activation. 

The latent variables of the feature vector are fed to independent GPs with constant mean and ARD-SE kernel, which produces a 20-dimensional latent state distribution $p(\mathbf{z}_t|\mathbf{x}_t)$. From the latent state distribution $p(\mathbf{z}_t|\mathbf{x}_t)$, we can sample the latent state vectors $\mathbf{z}_t$ using the reparametrization trick \cite{kingma2013auto}.

Similar to VAEs, the latent state vectors $\mathbf{z}_t$ are fed into the decoder $D$ to learn the reconstruction distribution $p(\hat{\mathbf{x}}_t|\mathbf{z}_t)$. A popular choice for $p(\hat{\mathbf{x}}_t|\mathbf{z}_t)$ is Gaussian with unit variance\cite{hafner2019learning, hafner2019dream, hafner2020mastering}. The decoder $D$ is parametrized by an NN composed of a linear fully-connected layer and 4 transpose convolutional layers with 32 filters each. The convolutional filters are of size $(3 \times 3)$ and shifted across the images with stride 1 (again, the last convolutional layer has stride 2). Batch normalization is used after the 2nd and 4th convolutional layers, and ELU activations are employed for all the layers except the last one. The outputs are the mean $\mu_{\hat{x}}$ and variance $\sigma^2_{\hat{x}_t}$ of $\mathcal{N}(\mu_{\hat{x}_t},\sigma^2_{\hat{x}_t})$. 

\subsection*{A.2 SVDKL Dynamical Model Architecture}\label{svdkl-fwd}
Given a sample $\mathbf{z}_t$ from the latent state distribution $p(\mathbf{z}_t|\mathbf{x}_t)$, we predict the evolution of the dynamical system forward in time with a control input $u$ using the SVDKL dynamical model $F$. The SVDKL dynamical model is composed of 3 fully-connected layers of size 512, 512, and 20, respectively, with ELU activations except the final layer with a linear activation. Analogously to the SVDKL encoder, the output features of the neural network are fed to 20 independent GPs to produce a 20-dimensional next state distribution $p(\mathbf{z}_{t+1}|\mathbf{z}_t, \mathbf{u}_t)$. Again, we sample the next latent states $\mathbf{z}_{t+1}$ using the reparametrization trick.

\subsection*{A.3 Pendulum Environment}\label{app_pendulum}
The pendulum environment used for collecting the data tuples is the {\fontfamily{cmtt}\selectfont Pendulum-v1} from Open-Gym \cite{brockman2016openai}.

\subsection*{A.4 KL Balancing}\label{app_kl_balance}

Similar to Dreamer-v2\cite{hafner2020mastering}, we employ the KL balancing. The method allows for balancing how much the prior is pulled towards the posterior and vice versa, and can be easily implemented as follows:
\begin{equation*}
\mathcal{L}_{KL} = \mathbb{E}_{\mathbf{x}_t, \mathbf{x}_{t+1} \sim \mathbf{X}, \mathbf{u}_t \sim \mathbf{U}}\Big[  \alpha \text{KL}\big[{\color{gray}\text{stop\_grad}(p(\mathbf{z}_{t+1}|\mathbf{x}_{t+1}))} ||  p(\mathbf{z}_{t+1}|\mathbf{z}_t,\mathbf{u}_t)\big] + (1-\alpha)  \text{KL}\big[p(\mathbf{z}_{t+1}|\mathbf{x}_{t+1}) ||  {\color{gray}\text{stop\_grad}(p(\mathbf{z}_{t+1}|\mathbf{z}_t,\mathbf{u}_t))}\big]\Big]\,,
\end{equation*}
where $\alpha$ is a hyperparameter balancing the contribution of the two terms of the KL divergence, and $\text{stop\_grad}$ is the function stopping the propagation of the gradients during the update step of the SVDKL parameters.

\subsection*{A.5 Hyperparameter Summary}

The hyperparameters are chosen via grid search among the values reported in Table \ref{tab1}. The final values used in the experiments are indicated in bold.
\begin{table}[h!]
\centering
\begin{tabular}{|c | c|} 
\hline
 \hline
 Hyperparameter & Value \\ [0.5ex] 
 \hline\hline
learning rate of NN  & $[1\mathrm{e}{-5}, 1\mathrm{e}{-4}, 2\mathrm{e}{-4}, \mathbf{3e-4}, 4\mathrm{e}{-4}, 5\mathrm{e}{-4}, 1\mathrm{e}{-3}, 1\mathrm{e}{-2}]$  \\ 
 \hline
learning rate of GP & $[1\mathrm{e}{-1}, \mathbf{1e-2}, 1\mathrm{e}{-3}]$\\
\hline
$L^2$ regularization coefficient & $[1\mathrm{e}{-1}, \mathbf{1\mathrm{e}{-2}}, 1\mathrm{e}{-3}, 1\mathrm{e}{-4},1\mathrm{e}{-5}]$ \\
\hline
$\alpha$  & $[0.8, \mathbf{0.9}, 1.0]$   \\
\hline
 $\beta$ & $[\mathbf{1.0}]$  \\
  \hline
  latent state dimension & $[5, 10, \mathbf{20}, 50]$ \\
 \hline
 number of inducing points & $[\mathbf{32}, 64]$ \\ 
 \hline
 \hline
\end{tabular}
\caption{Hyperparameters in the proposed SVDKL-based scheme. \textbf{Bold font} indicates the actual value used for generating the results.}
 \label{tab1}
\end{table}
Other parameters used in the experiments are listed in Table \ref{tab2}.
\begin{table}[h!] 
\centering
\begin{tabular}{|c | c|}
\hline
 \hline
 Other parameter & Value \\ [0.5ex] 
 \hline\hline
 image dimension & $84\times84\times3$ \\
  \hline
  measurement dimension & $84\times84\times3\times2$ \\
  \hline
  control input dimension & 1 \\
 \hline
 mass of the pendulum & 1 \\
 \hline
 length of the pendulum & 1 \\
 \hline
 $\sigma^2_x$ & $[0.1, 0.2, 0.3, 0.4,0.5,0.6, 0.7,0.8,0.9, 1.0]$  \\ 
 \hline
 $\sigma^2_u$ & $[0.2, 0.4, 0,5, 0.6, 0.8, 1.0, 1.5]$  \\ 
 \hline
 $\sigma^2_{dyn}$ & $[0.5, 1.0, 5.0, 10.0, 50.0, 100.0, 200.0]$  \\ 
 \hline
 \hline
\end{tabular}
\caption{Other parameters used in numerical experiments.}
 \label{tab2}
\end{table}

{\color{black}

\section*{Appendix B Comparison with Variational Autoencoder }

We compare the proposed SVDKL-based scheme with a VAE-based counterpart\cite{kingma2013auto} in the low-dimensional learning of state representation and latent forward model (for the pendulum) using high-dimensional noisy measurements.

\subsection*{B.1 Architecture}

For a fair comparison, the VAE- and SVDKL-based schemes have very similar model architectures.
We use the same encoding architecture (see Section \ref{svdkl-ae}), and the two models only differ in the last layer of encoder, i.e., the outputs of the VAE are the means and standard deviations of the Gaussian distributions for latent states.
Similarly, the NN-based latent forward models are formulated identically (see Section \ref{svdkl-fwd}), expect that the means and standard deviations of the next latent states are defined as outputs in the VAE-based scheme. 

\subsection*{B.2 Loss Function}

To train the VAE-based models, we employ a VAE loss $\mathcal{L}_{E}(\mathbf{\theta}_E, \mathbb{\theta}_D)$, a dynamical model loss $\mathcal{L}_{\text{F}}(\mathbf{\theta}_F)$, and the overall loss $\mathcal{L}_{REP}(\mathbf{\theta}_E, \mathbf{\theta}_F, \mathbb{\theta}_D)$ defined as their combination, all of which take the same form as their counterparts in the SVDKL-based scheme respectively, expect that there are no longer kernel hyperparameters to be determined. Because the VAE directly learns the mean and the standard deviation of the Gaussian distribution, we do not need to perform variational inference as in the case of the SVDKL-based models. 

\subsection*{B.3 Hyperparameters}

The hyperparameters for the VAE-based model training are set to the same values as in the SVDLK-based scheme, reported in Table \ref{tab3}.
\begin{table}[h!]
\centering
\begin{tabular}{|c | c|} 
\hline
 \hline
 Hyperparameter & Value \\ [0.5ex] 
 \hline\hline
learning rate of NN  & $3\mathrm{e}-4$  \\ 
 \hline
$L^2$ regularization coefficient & $1\mathrm{e}-2$ \\
\hline
$\alpha$  & $0.9$   \\
\hline
 $\beta$ & $1.0$  \\
  \hline
  latent state dimension & $20$ \\
 \hline
 \hline
\end{tabular}
\caption{Hyperparameters in the VAE-based scheme for comparative purposes.}
 \label{tab3}
\end{table}

\subsection*{B.4 Results}
To empirically demonstrate the advantages in using the SVDKL-based scheme over the VAE-based models for state estimation and denoising, we show the reconstructed images under different noise levels in Figure \ref{fig:reconstructions_VAEvSVDKL}. The SVDKL-based models, especially in the case of high measurement noise (e.g., $\sigma_u=0.7$ or $\sigma_x=1.0$), provide sharper reconstructions.
\begin{figure}[h!]
     \centering
     \begin{subfigure}[b]{0.33\textwidth}
         \centering
         \includegraphics[width=1.0\textwidth]{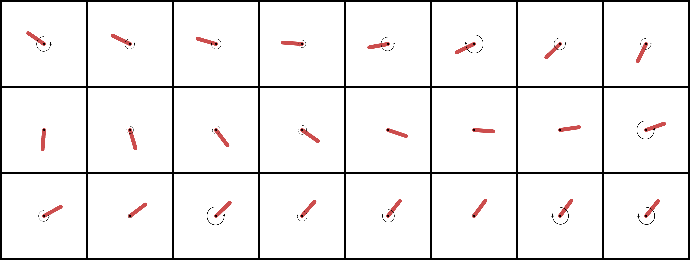}
         \caption{$\sigma_x=0.0, \sigma_u=0.7$}
         \label{fig:meas_noise=0.0}
     \end{subfigure}
     \hfill
     \begin{subfigure}[b]{0.33\textwidth}
         \centering
         \includegraphics[width=1.0\textwidth]{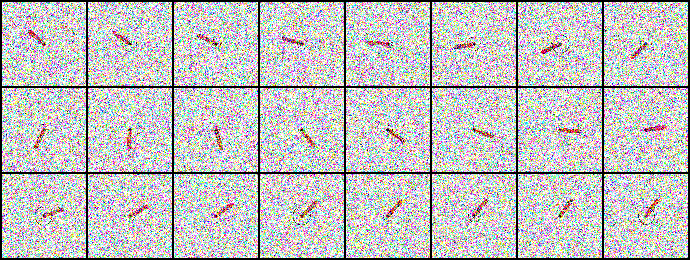}
         \caption{$\sigma_x=0.5, \sigma_u=0.5$}
         \label{fig:meas_noise=0.5}
     \end{subfigure}
     \hfill
    \begin{subfigure}[b]{0.33\textwidth}
         \centering
         \includegraphics[width=1.0\textwidth]{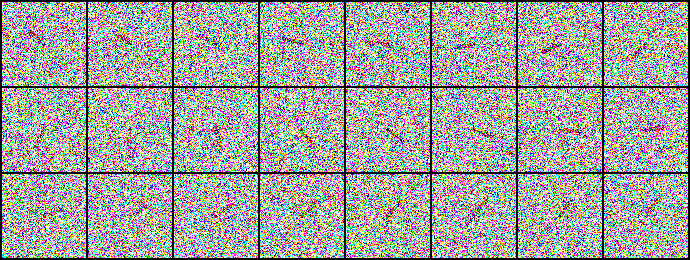}
         \caption{$\sigma_x=1.0, \sigma_u=0.0$}
         \label{fig:meas_noise=1.0}
     \end{subfigure}
     \begin{subfigure}[b]{0.33\textwidth}
         \centering
         \includegraphics[width=1.0\textwidth]{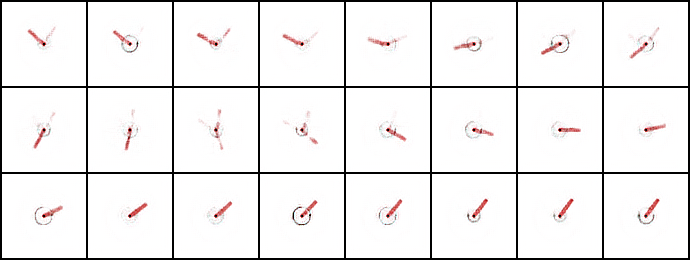}
         \caption{$\sigma_x=0.0, \sigma_u=0.7$}
         \label{fig:VAEnoise=0.0}
     \end{subfigure}
     \hfill
     \begin{subfigure}[b]{0.33\textwidth}
         \centering
         \includegraphics[width=1.0\textwidth]{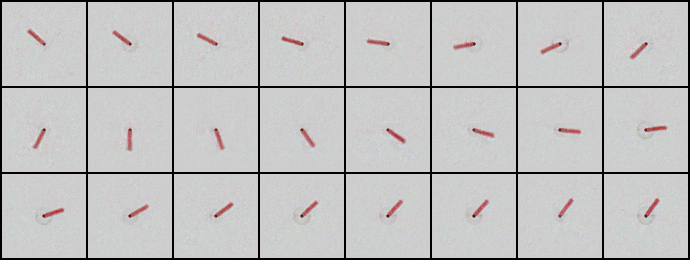}
        \caption{$\sigma_x=0.5, \sigma_u=0.5$}
         \label{fig:VAEnoise=0.5}
     \end{subfigure}
     \hfill
    \begin{subfigure}[b]{0.33\textwidth}
         \centering
         \includegraphics[width=1.0\textwidth]{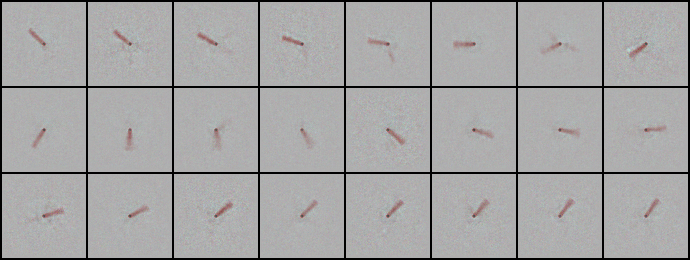}
         \caption{$\sigma_x=1.0, \sigma_u=0.0$}
         \label{fig:VAEnoise=1.0}
     \end{subfigure}
\begin{subfigure}[b]{0.33\textwidth}
         \centering
         \includegraphics[width=1.0\textwidth]{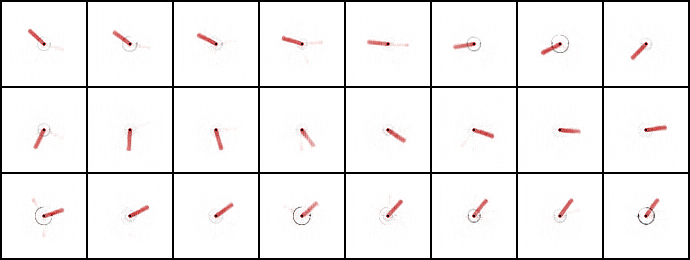}
         \caption{$\sigma_x=0.0, \sigma_u=0.7$}
         \label{fig:SVDKLnoise=0.0}
     \end{subfigure}
     \hfill
     \begin{subfigure}[b]{0.33\textwidth}
         \centering
         \includegraphics[width=1.0\textwidth]{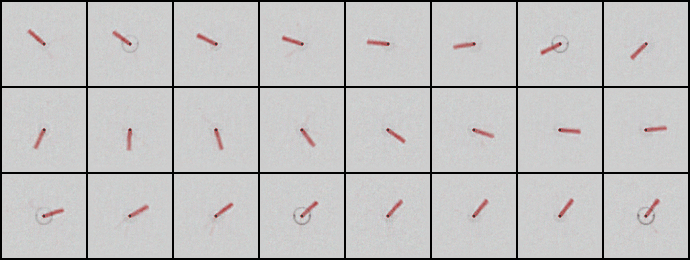}
         \caption{$\sigma_x=0.5, \sigma_u=0.5$}
         \label{fig:SVDKLnoise=0.5}
     \end{subfigure}
     \hfill
    \begin{subfigure}[b]{0.33\textwidth}
         \centering
         \includegraphics[width=1.0\textwidth]{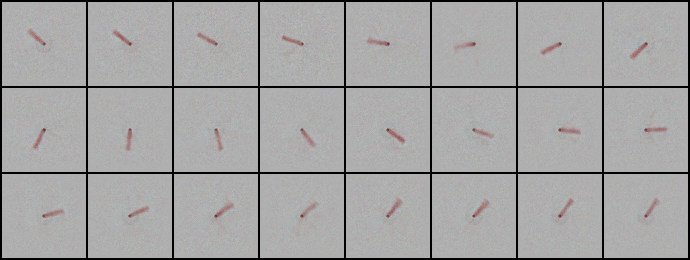}
         \caption{$\sigma_x=1.0, \sigma_u=0.0$}
         \label{fig:SVDKLnoise=1.0}
     \end{subfigure}
\caption{Measurements (Figure \ref{fig:meas_noise=0.0}-\ref{fig:meas_noise=1.0}) with different noise levels and the corresponding reconstructions through the VAE-based scheme (Figure \ref{fig:VAEnoise=0.0}-\ref{fig:VAEnoise=1.0}) and the SVDKL-based scheme (Figure \ref{fig:SVDKLnoise=0.0}-\ref{fig:SVDKLnoise=1.0}).}
        \label{fig:reconstructions_VAEvSVDKL}
\end{figure}

}